\documentclass{article}

\PassOptionsToPackage{table, usenames, dvipsnames}{xcolor}

\usepackage{pdfpages} 
\usepackage{pgffor} 

\makeatletter
\AtBeginDocument{\let\LS@rot\@undefined}
\makeatother

\def\supplementfilename{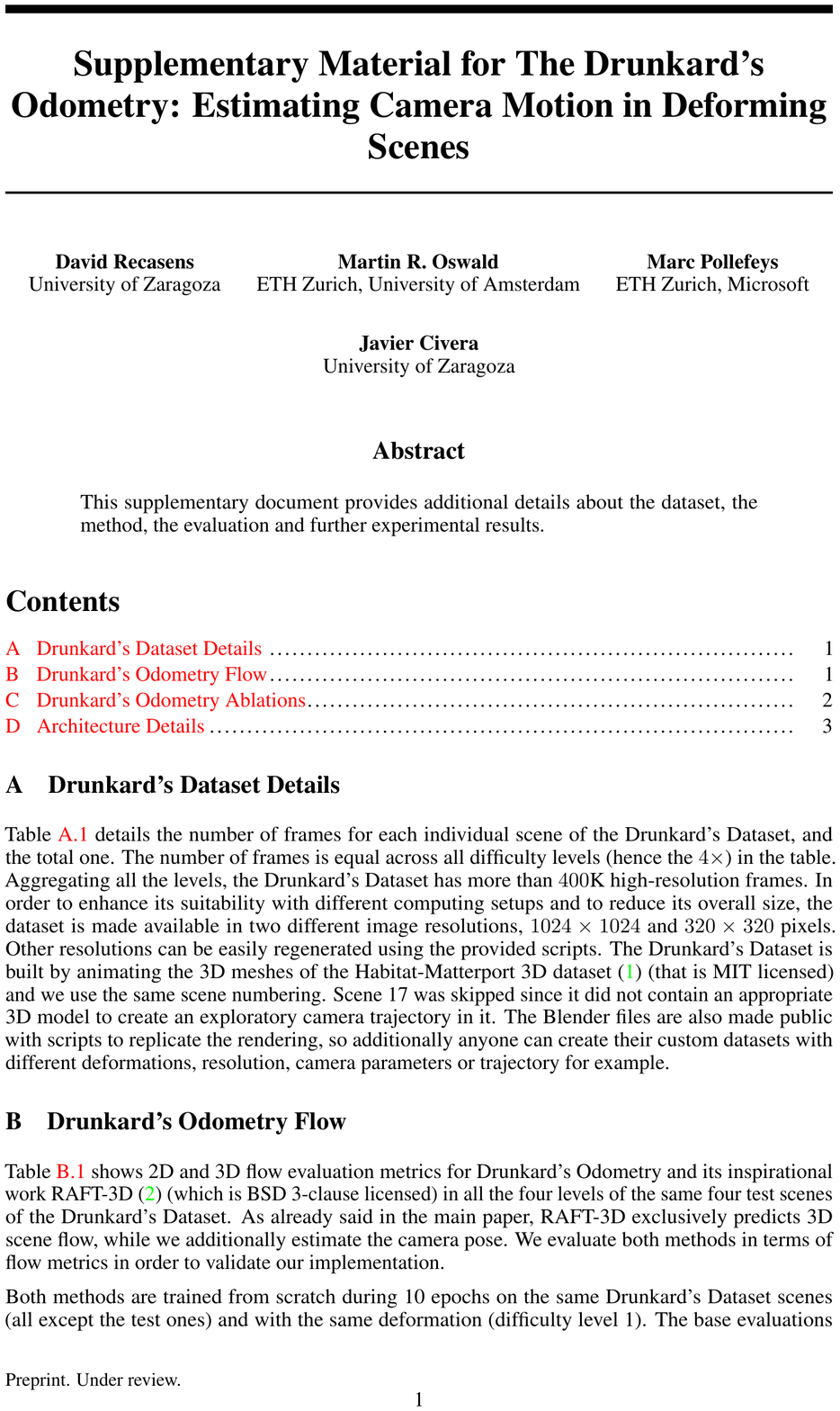}

\pdfximage{\supplementfilename}
\def\numbersupplementpages{\the\pdflastximagepages}

\newif\ifarXiv
\arXivtrue 

     \usepackage[nonatbib,preprint]{neurips_data_2023}

\usepackage[utf8]{inputenc} 
\usepackage[T1]{fontenc}    
\usepackage{url}            
\usepackage{booktabs}       
\usepackage{amsfonts}       
\usepackage{nicefrac}       
\usepackage{microtype}      
\usepackage[table, usenames, dvipsnames]{xcolor}         
\usepackage{times}
\usepackage{epsfig}
\usepackage{graphicx}
\usepackage{amsmath}
\usepackage{amssymb}
\usepackage{comment}
\usepackage{media9}
\usepackage{subcaption}
\usepackage{footmisc}
\usepackage{wrapfig}

\usepackage{multirow}
\usepackage{makecell}
\usepackage{colortbl}
\usepackage{pifont}
\usepackage{balance}
\usepackage{arydshln}  

\usepackage[capbesideposition=outside,capbesidesep=quad]{floatrow}		

\usepackage{xspace} 
\usepackage{xfrac}

\def\etal{\emph{et al.}\xspace}

\newcommand{\cmark}{\textcolor{OliveGreen}{\ding{51}}}%
\newcommand{\xmark}{\textcolor{red}{\ding{55}}}%

\DeclareMathAlphabet{\pazocal}{OMS}{zplm}{m}{n}
\newcommand{\Lb}{\pazocal{L}}

\newcommand{\boldparagraph}[1]{\vspace{0.0em}\noindent{\bf #1}}

\colorlet{colorFst}{Green!25}       
\colorlet{colorSnd}{SpringGreen!45} 
\colorlet{colorTrd}{Yellow!30}      
\colorlet{colorLow}{darkgray!30}    
\newcommand{\fs}{\cellcolor{colorFst}\bf}   
\newcommand{\nd}{\cellcolor{colorSnd}}      
\newcommand{\rd}{\cellcolor{colorTrd}}      

\usepackage[pagebackref=true,breaklinks=true,colorlinks,bookmarks=false]{hyperref}

\title{The Drunkard’s Odometry: Estimating Camera Motion in Deforming Scenes}

\author{%
  David Recasens \\
  University of Zaragoza\\  
  \And
  Martin R.~Oswald \\
  ETH Zurich, University of Amsterdam \\
  \And
  Marc Pollefeys \\
  ETH Zurich, Microsoft \\
  \And
  Javier Civera \\
  University of Zaragoza \\  
}

\begin{document}

\maketitle

\begin{abstract}
    Estimating camera motion in deformable scenes poses a complex and open research challenge. 
    Most existing non-rigid structure from motion techniques assume to observe also static scene parts besides deforming scene parts in order to establish an anchoring reference. However, this assumption does not hold true in certain relevant application cases such as endoscopies. Deformable odometry and SLAM pipelines, which tackle the most challenging scenario of exploratory trajectories, suffer from a lack of robustness and proper quantitative evaluation methodologies. To tackle this issue with a common benchmark, we introduce the Drunkard's Dataset, a challenging collection of synthetic data targeting visual navigation and reconstruction in deformable environments. This dataset is the first large set of exploratory camera trajectories with ground truth inside 3D scenes where every surface exhibits non-rigid deformations over time. Simulations in realistic 3D buildings lets us obtain a vast amount of data and ground truth labels, including camera poses, RGB images and depth, optical flow and normal maps at high resolution and quality.
    We further present a novel deformable odometry method, dubbed the Drunkard’s Odometry, which decomposes optical flow estimates into rigid-body camera motion and non-rigid scene deformations.
    In order to validate our data, our work contains an evaluation of several baselines as well as a novel tracking error metric which does not require ground truth data.
    Dataset and code: \url{https://davidrecasens.github.io/TheDrunkard'sOdometry/}
\end{abstract}


\section{Introduction} \label{sec:intro}
%
\begin{figure}
  \centering
  \footnotesize
  \setlength{\tabcolsep}{0pt}
  \setlength\extrarowheight{55pt}
  \begin{tabular}{cc}
    \rotatebox[origin=c]{90}{\hspace{22pt}camera view} &
    \multirow{2}{*}[0pt]{
    \includemedia[
      width=0.96\linewidth,    
      activate=pageopen,
      addresource=Images/teaserLow2.mp4,
      flashvars={
        source=Images/teaserLow2.mp4
      }
    ]{\includegraphics{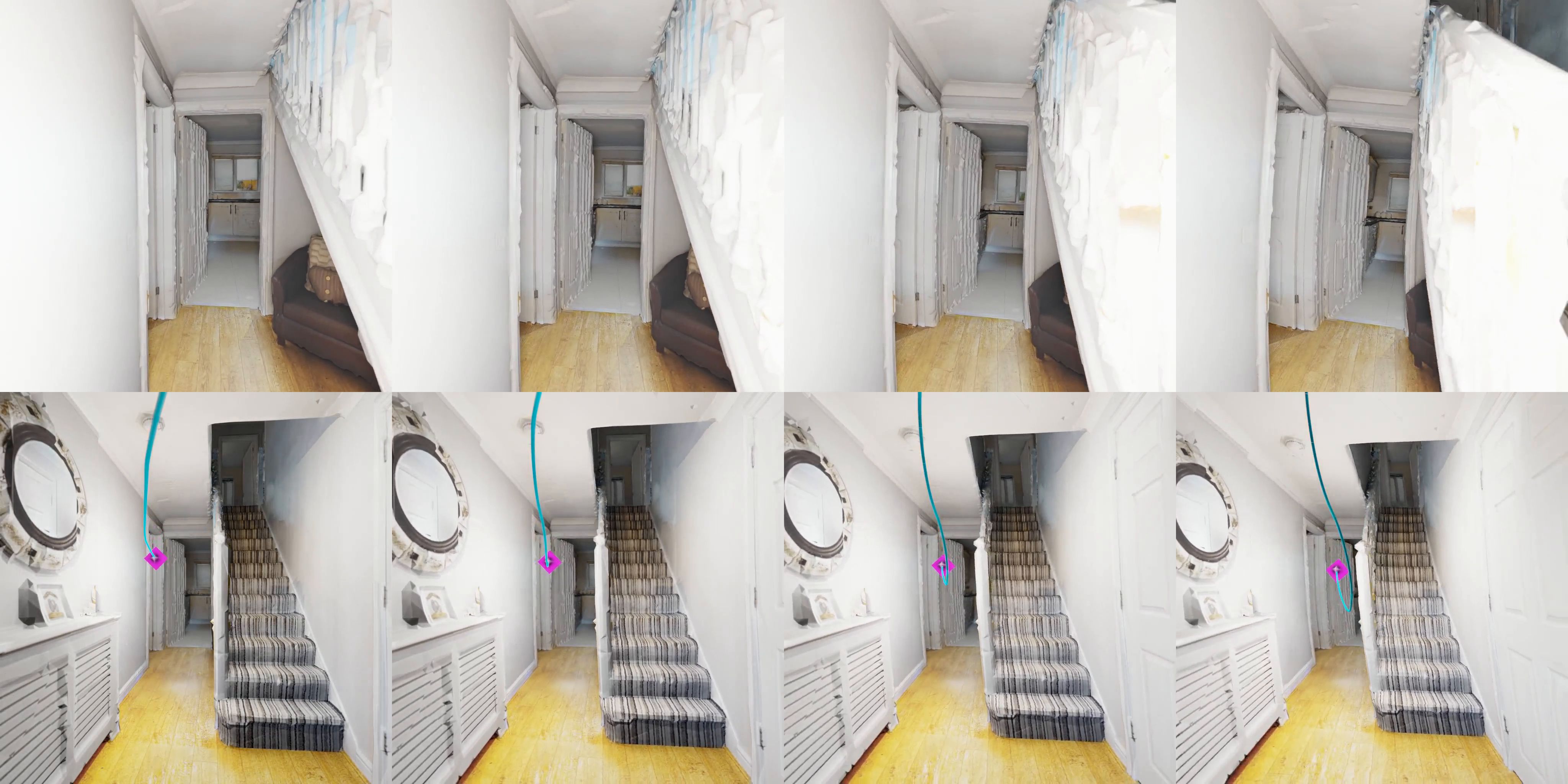}}{VPlayer.swf}
    } \\
    \rotatebox[origin=c]{90}{\hspace{20pt}3rd person view}
  \end{tabular}
  \setlength{\tabcolsep}{35pt}
  \setlength\extrarowheight{0pt}
  \begin{tabular}{cccc}
    ~~~~~Level 0 & Level 1 & Level 2 & Level 3 \\[-5pt]
  \end{tabular}
  \caption{\textbf{Sample scene of the Drunkard’s Dataset.} The dataset provides various levels of scene deformation. \textbf{Top row:} Sample frames from scene 0 over all difficulty levels $0-3$. \textbf{Bottom row:} External views showing the ground truth camera trajectory in green and the camera frame in purple. With increasing deformation level the camera motion is more abrupt.
  See animations with \emph{Adobe Reader}, \emph{KDE Okular} or \href{https://youtu.be/wL8JDg6bemg}{YouTube}.
  }
  \label{fig:teaser}
\end{figure}

Deformable scenes are among the most challenging cases for visual navigation and multi-view reconstruction. They may also be among the ones with most relevant potential applications, ranging from the reconstruction of deforming objects~\cite{badias2021morph}, faces~\cite{li2021topologically}, hands~\cite{qian2020html}, human bodies~\cite{wang2021deep} (or animals'~\cite{li2021coarse}), clothing~\cite{su2020mulaycap}) or the body interior for medical applications~\cite{lamarca2020defslam,sengupta2021colonoscopic,recasens2021endo,shao2022self}.
Among all potential applications for mapping and navigation in non-rigid scenes, medical ones stand out as very different from the rest and are the target of our work. In certain medical procedures, such as endoscopies, a camera navigates inside the human body, performing \emph{exploratory trajectories} that extend far beyond its field of view. For the rest of the applications mentioned, the camera remains nearly or fully stationary and most views have a high degree of overlap. The field of Non-Rigid Structure from Motion (NRSfM) has experienced significant progress in the last decades, 
e.g. \cite{rabaud2008re,gotardo2011non,lee2013procrustean,garg2013dense,dai2014simple,agudo2014good,kong2019deep,bozic2020deepdeform,yang2021lasr,tretschk2021non,pumarola2021d,potje2021extracting,parashar2021robust,weng2022humannerf}. However, most of them address small-scale reconstructions and are of limited use in medical applications. For the few exceptions covering SLAM~\cite{cadena2016past} in deformable scenes (e.g., \cite{lamarca2020defslam,sengupta2021colonoscopic,shao2022self,gomez2021sd}), there are no clear benchmarks nor datasets to support and track progress in the field.

Our main contribution in this work is motivated by this need for benchmarking exploratory camera motion in deforming scenes, a problem we will refer to as the ``Drunkard's odometry''. Publicly available datasets imaging deforming scenes do not cover exploratory trajectories, and the ones that do cover them do not have ground truth geometric annotations. Our proposal is a synthetic dataset that we denote as the ``Drunkard's Dataset'', containing a set of high-resolution RGB images, ground truth depth, optical flow, normal maps and camera pose trajectories in synthetic but texture-realistic deforming scenes. In order to generate a sufficient amount and variety of data, we imported the real-world scanned indoor 3D models of the Habitat-Matterport 3D dataset~\cite{ramakrishnan2021habitat}, added dynamic deformations, and generated trajectories within them. 
Figure~\ref{fig:teaser} shows several data samples with the camera trajectory that advances through time. As the original 3D models are real-world scans, camera trajectory and depth are in consistent metric scale along scenes. To make it a scalable benchmark dataset, every scene was recorded under four different levels of reconstruction difficulty, increasing the deformation and camera trajectory perturbations (observe them in Figure~\ref{fig:teaser}). The Drunkard’s Dataset is unique in its kind, providing large-scale data in deformable scenes, which will enable both, benchmarking non-rigid navigation and reconstruction methods, as well as sufficient data to evaluate the potential benefits of deep learning. Please visit our project website for further details and access to the dataset and source code.

Scientific progress is in many occasions based on well-established and public benchmarks. This has been the specific case in computer vision research in the last decades, up to the point that the existence of some benchmarks is highly correlated with scientific progress in the field. Data repositories are also essential nowadays not only to benchmark different methods, but also to train deep neural networks. However, collecting large amounts of data with non-rigid content is challenging due to difficulties in annotation, as argued by Li et al.~\cite{li2021topologically}. Indeed, some very popular datasets are synthetic~\cite{butler2012naturalistic,mayer2016large}. 

Capturing large-scale data in sufficiently large non-rigid spaces, in order to benchmark odometry/SLAM methods, is even more challenging. Having ground truth annotations, in particular in the medical domain, hugely increases the challenge. The most popular medical datasets~\cite{stoyanov2005soft,stoyanov2012stereoscopic,jensen2021benchmark,azagra2022endomapper} are small, lack geometric ground truth, or both. As a result, comparisons between methods are very often inaccurate, inconclusive or questionable. This motivates our work.

A second contribution of this paper is the Drunkard's Odometry, a flow-based odometry method for camera motion estimation from RGB-D sequences in deformable scenes. Our method is inspired by the pose estimation of DROID-SLAM~\cite{teed2021droid} and by the 3D scene flow prediction of RAFT-3D~\cite{teed2021raft}. Our novelty with respect to both is that our architecture models potential scene deformations. 

As a third contribution, in order to validate methods also in non-annotated data, we present a novel metric, the Absolute Palindrome Trajectory Error ($\mathrm{APTE}$). Our novel metric is based on running odometry methods forwards and backwards in a sequence of images, and comparing the pose errors between the first and last frames through different loop lengths. This metric may be useful for validating methods in realistic setups, but it is limited to relative errors between two poses and hence it is not as informative as metrics using ground truth labels. For this reason, we believe that $\mathrm{APTE}$ in real sequences should not be trusted on its own, but as a complement to a larger and more informative set of metrics in our simulated Drundard's Dataset.

\section{Related Work}  \label{sec:related}
%
\begin{wraptable}[12]{r}{8.1cm}
  \vspace{-12pt}
  \centering
  \scriptsize
  \setlength{\tabcolsep}{2pt}
  \renewcommand{\arraystretch}{1.15}
  \begin{tabular}{lccccc}
     \toprule
     Dataset & Sim/Real & GT & \makecell{Explo-\\ratory} & \#Frames & Domain \\ 
     \midrule
     De Aguiar~\etal~\cite{de2008performance} & Real & \cmark & \xmark & $\sim6$K & Human bodies   \\ 
     FAUST~\cite{bogo2014faust} & Real & \cmark & \xmark & $\sim9$K & Human bodies   \\ 
     NRSfM Challenge~\cite{jensen2021benchmark} & Real & \cmark & \xmark &- & Objects   \\ 
     DeformingThings4D~\cite{li20214dcomplete} & Sim & \cmark & \xmark & $\sim122$K & Humans \& animals \\
     ToFu~\cite{li2021topologically} & Sim & \cmark &\xmark& $\sim20$K & Faces \\ 
     EndoMapper~\cite{azagra2022endomapper} & Real & \xmark &\cmark& $\sim4$M & Colonoscopies \\
     Hamlyn~\cite{mountney2010three,stoyanov2005soft,stoyanov2010real,pratt2010dynamic} & Real & \xmark &\cmark& $\sim93$K & Laparoscopies \\
     \textbf{Drunkard's (ours)} & Sim & \cmark &\cmark& $\sim416$K & Indoor scenes \\ 
     \bottomrule     
  \end{tabular}
  \vspace{-4pt}
  \caption{\textbf{Overview of existing non-rigid dataset specifications} compared to our Drunkard's dataset.}
  \label{tab:related}
\end{wraptable}
Structure from Motion, visual odometry and visual SLAM methods for rigid scenes are commonly evaluated in a well established set of publicly available datasets~\cite{sturm2012benchmark,geiger2013vision,wilson2014robust,burri2016euroc,dai2017scannet,schubert2018tum,schops2019bad}, with ground truth geometric annotations and under the same metrics, which facilitates comparisons between them and progress in the field. Large-scale annotations for geometric ground truth are quite challenging in real scenarios, and still contain small errors due to the accuracy of the equipment (GPS or motion capture systems) used. An accurate approximation of the depth and camera pose could be recorded with sensors, but could not get such a good quality of optical flow and normal maps.
In real-world applications that involve significant deformations, such as endoscopies, these sensors cannot be even equipped. While we could attempt to simulate a small-scale scenario using moving blankets or bouncy castles, it would fall short in terms of capturing a wide range of textures, deformations, and camera trajectories. Virtual environments, on the other hand, offer a superior solution, providing a rich and diverse representation of the complex conditions found in real-world settings.
Synthetic datasets solve these issues at the price of the sim2real gap, and are common for benchmarking methods on navigation and reconstruction on rigid scenes~\cite{handa2014benchmark,straub2019replica,wang2020tartanair,minoda2021viode, ramakrishnan2021habitat}. For similar reasons, synthetic datasets are widely used in other computer vision tasks, such as stereo and flow~\cite{baker2011database,butler2012naturalistic,dosovitskiy2015flownet,mayer2016large,ilg2018occlusions,mehl2023spring}, depth~\cite{carlucci2017deep,planche2017depthsynth}, object recognition and segmentation~\cite{chang2015shapenet}, object pose estimation and tracking~\cite{gaidon2016virtual}, and scene segmentation,  understanding and reasoning~\cite{mccormac2017scenenet,johnson2017clevr,roberts2021hypersim}. For additional insights and references on the use of synthetic data in deep learning, the reader is referred to the excellent survey by Nikolenko~\cite{nikolenko2019synthetic}.

Note that, in most of the datasets cited, in particular those for faces and humans bodies, the camera is almost stationary. For the few datasets where the camera moves sufficiently, exploring new areas, the covered region is in any case reduced, deformations are small or there is absence of geometric ground truth (see Table~\ref{tab:related}). This poses difficulties for benchmarking methods for odometry and SLAM targeting non-rigid environments, such as~\cite{lamarca2020defslam,recasens2021endo}. 
Odometry and SLAM methods for deformable scenes are scarce in the literature, being \cite{song2018mis,lamarca2020defslam,sengupta2021colonoscopic,recasens2021endo,gomez2021sd,shao2022self,lamarca2022direct,rodriguez2022tracking} the most representative ones. However, they are all based on feature matching or direct tracking, which make them unstable in challenging sequences. Differently, our Drunkard's Odometry is based on scene flow, which makes it significantly more robust.

\section{The Drunkard's Dataset}  \label{sec:dataset}
%
\begin{figure*}[htb!]
    \centering
    \includegraphics[width=1.\linewidth]{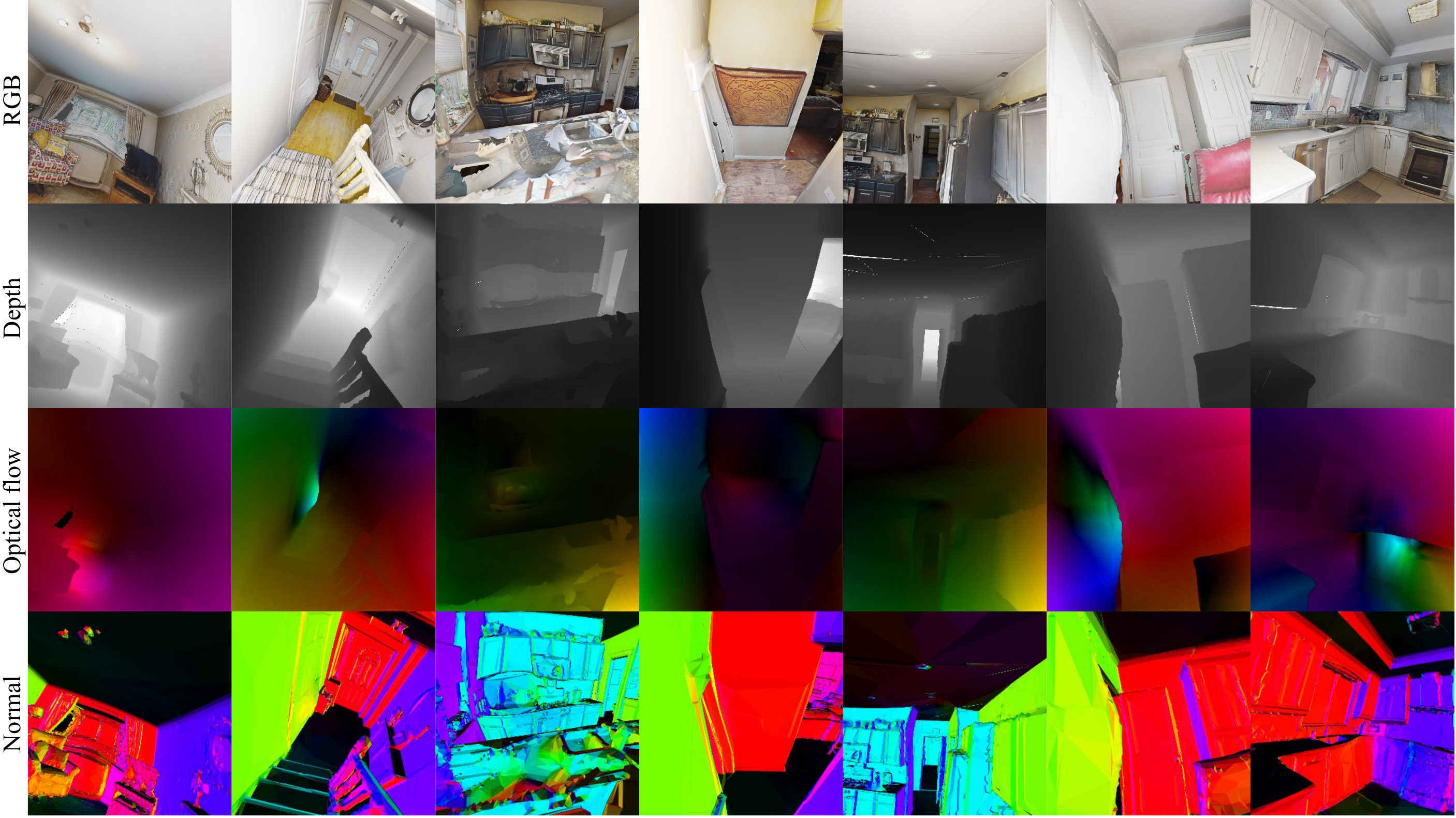}\\[-6pt]
    \caption{\textbf{Sample images of the Drunkard's Dataset} with their corresponding depth, optical flow and normal maps. Non-rigid deformations are simulated by smooth deformations of all scene parts.}
    \label{fig:sampled_images}    
\end{figure*}
The Drunkard's Dataset is a publicly available set of $19$ different camera trajectory recordings of $19$ different deforming indoor scenes, where each one has been recorded four times, one for each difficulty level. For each level, we generated over $100$K frames and recorded camera poses, RGB images, depths, optical flow and normal maps at $1920 \times 1920$, being the camera poses and the depth in real-world metric scale consistent throughout different scenes. Find sample images in Figure~\ref{fig:sampled_images}.

\begin{table}    
    \centering
    \scriptsize
    \setlength{\tabcolsep}{5pt}
    \renewcommand{\arraystretch}{0.8}
    \floatbox[\capbeside]{table}[0.4\columnwidth]%
    {\caption{\textbf{Overview of the four difficulty levels.} Deformation and camera trajectory noises increases with higher levels. At level 0 there is neither deformation nor camera noise, i.e.\ it represents a rigid scene and a smooth motion.} 
    \label{tab:table_drunk_levels}}{
    \begin{tabular}{ccc}
        \toprule
        Difficulty level & Deformations & Trajectory noise \\
        \midrule
        Level 0 & 0  & 0    \\
        \rd Level 1 & \rd Low     & \rd Low    \\
        \nd Level 2 & \nd Medium  & \nd Medium \\
        \fs Level 3 & \fs High    & \fs High   \\     
        \bottomrule
    \end{tabular}%
    }    
\end{table}

We used Blender~\cite{blender} to render the deformations of the $19$ real-world scanned indoor 3D models of the Habitat-Matterport 3D dataset~\cite{ramakrishnan2021habitat}.
We manually designed camera trajectories such that every room in each building is visited once and the camera trajectory ends at the starting point, except for scenes $4$, $9$ and $14$, in which the camera traverses the building three times, but in each loop visiting the rooms in different order. The Blender files are publicly available, along with the scripts we used for generating them, so that anyone can render in-house versions of the Drunkard's Dataset scenes modifying parameters of deformations, pose trajectory, resolution or camera type among others.

As Table~\ref{tab:table_drunk_levels} details, Level 0 stands for zero deformation and camera noise, resulting in a rigid scene and a smooth camera motion, well suited for rigid SfM/SLAM methods. 
The following levels have an increasing degree of deformation and trajectory noise. Having four levels of difficulty allows to benchmark both rigid and non-rigid methods in a graduated manner.

\section{The Drunkard's Odometry}  \label{sec:drunkSLAM}
%
Given a pair of RGB-D images $\{\mathbf{I}_i, \bar{\mathbf{Z}}_i\}$, $\mathbf{I}_i \in \mathbb{R}^{w \times h \times 3}$ and $\bar{\mathbf{Z}}_i \in \mathbb{R}^{w \times h}$, $i \in \{1, 2 \}$, our Drunkard's Odometry estimates a dense scene flow $\mathbf{T}^k \in \ensuremath{\mathrm{SE}(3)^{w \times h}}$ between them, and the relative camera motion $\mathbf{T}_c^k \in \ensuremath{\mathrm{SE}(3)}$ iteratively ($k \in \{1, \dots, M\}$ stands for the iterative block step). 
$\mathbf{T}^k$ contains pixel-wise rigid-body transformations that ideally, i.e.\ in absence of noise, maps every 3D point $\mathbf{P}_j = \pi^{-1}\left( \mathbf{u}_j, \bar{z}_j \right)$ --corresponding to pixel $j \in \mathbf{I}_1$, back-projected from the pixel with image coordinates $\mathbf{u}_j \in \Omega_1$ and its sensor depth $\bar{z}_j\equiv \bar{\mathbf{Z}}_1[\mathbf{u}_j]$-- to its ground truth equivalent 3D point $\bar{\mathbf{P}}_{j^\prime} = \pi^{-1}\left( \bar{\mathbf{u}}_{j^\prime}, \bar{z}_{j^\prime} \right)$ back-projected from the true associated pixel with image coordinates $\bar{\mathbf{u}}_{j^\prime} \in \Omega_2$ and sensor depth $\bar{z}_{j^\prime} \equiv \bar{\mathbf{Z}}_2[\bar{\mathbf{u}}_{j^\prime}]$. $\pi^{-1}\left( \mathbf{u}, z \right)$ stands for the inverse projection model. As the estimated scene flow $\mathbf{T}^k$ might be affected by noise, we will denote as $\mathbf{P}_{j\prime}^k = \mathbf{T}^k\left[\mathbf{u}_j\right] \mathbf{P}_j$
the transformation of $\mathbf{P}_j$ to the local frame of $\mathbf{I}_2$, that in general will not coincide the true corresponding point $\bar{\mathbf{P}}_{j^\prime}$. $\mathbf{T}^k\left[\mathbf{u}_j\right]$ is composed by the rigid transformation coming from the camera motion $\mathbf{T}_c^k$ and the one coming from the non-rigid surface deformations $\mathbf{T}_{d}^k\left[\mathbf{u}_j\right] \in \mathrm{SE}(3)$, so that $\mathbf{T}^k \left[\mathbf{u}_j\right]= \mathbf{T}_c^k  \mathbf{T}_{d}^k\left[\mathbf{u}_j\right]$. If the scene is rigid $\mathbf{P}_{j^\prime}^k = \mathbf{T}_c^k\mathbf{P}_j^k$, and if it is deforming $\mathbf{P}_{j^\prime}^k = \mathbf{T}_c^k\mathbf{T}_d^k\left[\mathbf{u}_j\right]\mathbf{P}_j^k$. 

\begin{figure*}[ht!]
    \centering
    \includegraphics[width=1\linewidth]{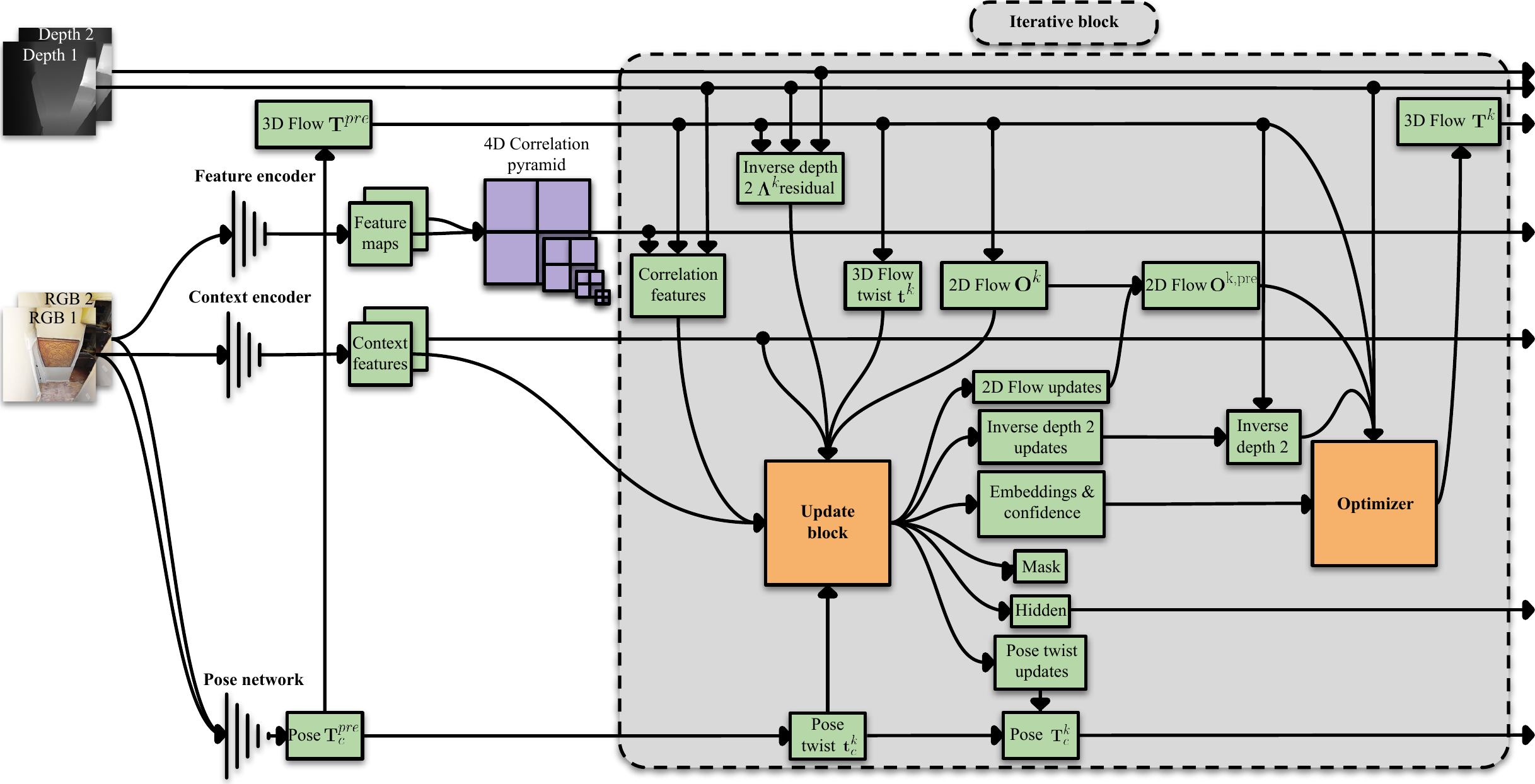}
    \caption{\textbf{Overview of the Drunkard's Odometry architecture.} 
    }
    \label{fig:overview}
\end{figure*}

Our iterative pose estimation model, shown in Figure~\ref{fig:overview}, is based on the foundations of the 3D flow estimation architecture of RAFT-3D~\cite{teed2021raft}, which do not have pose estimation capabilities at all. Firstly, a pose regression network encodes both color images $\mathbf{I}_1$ and $\mathbf{I}_2$ and outputs a initial pre-estimate for the camera motion $\mathbf{T}_c^{\text{pre}}$. This rigid transformation is used for initialization, so that for any $\mathbf{u}_j$ pixel of camera 1  $\mathbf{T}^{\text{pre}}\left[\mathbf{u}_j\right] = \mathbf{T}_c^{\text{pre}}$. Additionally, we encode both images $\mathbf{I}_1$ and $\mathbf{I}_2$ in two feature maps, which we use to build a 4D correlation pyramid between the features of all pixel pairs at four different scales, each scale halving the resolution of the previous one. From $\mathbf{T}^k$ and  $\bar{\mathbf{Z}}_1$ at each iteration $k$, we can obtain dense 2D pixel correspondences $\mathbf{u}_j \rightarrow \mathbf{u}_{j^\prime}$
\begin{equation}
   {\mathbf{u}_{j^\prime}^k = \pi \left( \pi^{-1}\left(\mathbf{u}_j, \bar{z}_j \right), \mathbf{T}^k\left[\mathbf{u}_j\right] \right)} \enspace,
\label{eq:1}
\end{equation}
where $\pi \left( \mathbf{P}, \mathbf{T}\left[\mathbf{u}\right] \right)$ stands for the pinhole camera projection model. At the beginning of each iteration, these correspondences are used to sample the correlation features from the fixed 4D correlation pyramid and it is one of the inputs of the update block. The estimated optical flow $\mathbf{O}^k \in \mathbb{R}^{w\times h\times 2}$ coming from these correspondences is also an input for the update block, being $\mathbf{O}^k[\mathbf{u}_j] = \mathbf{u}_{j^\prime}^k-\mathbf{u}_j$. 

The update block takes also as input an inverse depth residual obtained from the difference between the estimated inverse depth map $\mathbf{\Lambda}_2^k \in \mathbb{R}^{w \times h}$ and the sensor inverse depth $\bar{\mathbf{\Lambda}}_2\in \mathbb{R}^{w \times h}$, $\bar{\mathbf{\Lambda}}_2 \left[ \mathbf{u}^k_{j^\prime} \right] = \sfrac{1}{\bar{\mathbf{Z}}_2\left[ \mathbf{u}^k_{j^\prime} \right]}, \ \forall \mathbf{u}_{j^\prime}^k \in \Omega_2$. Values for $\mathbf{\Lambda}_2^k$ are interpolated from the grid defined by $\lambda_{2,j^\prime}^k = \mathbf{\Lambda}_2^k[\mathbf{u}_{j^\prime}^k] = \mathbf{e}_3^\top \mathbf{P}_{j^\prime}^k$.  $\mathbf{e}_3 = \left( 0 \ 0 \ 1 \right)^\top$ acts as selector of the third dimension of the 3D point.

A pair of context features that extracts semantic and contextual information from image 1 are also pre-rendered before entering the iterative process using a context encoder network and given to the update block. One context feature map is kept fix during all iterations and another is used as initialization of the hidden state of the General Recurrent Unit (GRU)~\cite{cho2014properties}, which is at the heart of the update block (see the network fine-grained details in the supplementary material). 

At each iteration $k$, the estimates for $\mathbf{T}^k$ and $\mathbf{T}_c^k$ are mapped to the Lie algebra with a logarithm map to result in a twist field $\mathbf{t}^k = \log_{\ensuremath{\mathrm{SE}(3)}}(\mathbf{T}^k)$ and $\mathbf{t}_c^k = \log_{\ensuremath{\mathrm{SE}(3)}}(\mathbf{T}_c^k)$ before being given to the update block. The update operator outputs a set of updates for the optical flow $\mathbf{O}^k$, for the twist camera pose $\mathbf{t}_c^k$ and for the inverse depth $\mathbf{\Lambda}_2^k$, the hidden state of the recurrent network, a mask and a set of rigid-motion embeddings and confidence maps. These last two maps, next to the updated estimate of $\mathbf{\Lambda}_2^k$ and $\mathbf{O}^k$, and $\mathbf{Z}_1$ are used by the least-squares optimizer block to update $\mathbf{T}^k$ (see details in \cite{teed2021raft} for this update). Scene flow is updated as $\mathbf{T}^{k+1} = \mathbf{T}^k \exp_{\ensuremath{\mathrm{SE}(3)}}(\mathbf{t}^k)$.

Internally the network works at ${1/8}^{th}$ of the original resolution, and the estimated mask at each iteration of the update block is used to perform a convex upsampling to the original resolution of $\mathbf{T}^k$ and intermediate updated 2D flow $\mathbf{O}^\text{k,pre} \in \mathbb{R}^{w\times h\times 2}$ by the update block. 

The supervision comes from comparing the estimated optical flow $\mathbf{O}^k$ obtained from  $\mathbf{T}^k$ and $\bar{\mathbf{Z}}_1$ with Eq.~\ref{eq:1} and the intermediate pre-estimated $\mathbf{O}^\text{k,pre}$ with the ground truth optical flow $\bar{\mathbf{O}} \in \mathbb{R}^{w\times h\times 2}$, the inverse depth error between $\mathbf{\Lambda}_2$ and $\bar{\mathbf{\Lambda}}_2$, the relative camera pose error of $\mathbf{T}_c$ and the initial guess pre-estimated by the pose network $\mathbf{T}_c^\text{pre}$ against $\bar{\mathbf{T}}_c$. The total loss results in
\begin{equation}
   \Lb = \Lb_{\text{pose}}^\text{pre} + \sum_{k=1}^{M} \gamma^{M-k} \left( \Lb_\text{flow}^k + \Lb_\text{depth}^k + \Lb_\text{pose}^k     \right) \enspace, 
\label{eq:3}
\end{equation}
with $\Lb_\text{flow}^k = \sum_{\mathbf{u}_j} \left( \lVert \mathbf{O}^k - \bar{\mathbf{O}} \rVert_1 + w_1 \lVert \mathbf{O}^\text{k,pre} - \bar{\mathbf{O}} \rVert_1 \right)$ being the optical flow loss term, $\Lb_\text{depth}^k = w_2 \sum_{\mathbf{u}_j} \lVert \mathbf{\Lambda}_2^k[\mathbf{u}_j] - \bar{\mathbf{\Lambda}}_2[\mathbf{u}_j] \rVert_1$ the inverse depth loss term, and $\Lb_{\text{pose}} = w_3 \lVert \log_{\ensuremath{\mathrm{SE}(3)}}\left(\mathbf{T}_c \bar{\mathbf{T}}_c^{-1} \right) \rVert_1 $ and $\Lb_{\text{pose}}^{\text{pre}} = w_4 \lVert \log_{\ensuremath{\mathrm{SE}(3)}}\left(\mathbf{T}^{\text{pre}}_c \bar{\mathbf{T}}_c^{-1} \right) \rVert_1$ the relative camera pose loss terms. $w_l$ stands for the relative weight of the $l^{th}$ loss term and $\gamma$ weights each loop.

\section{Experiments}  \label{sec:experiments}
%
In this section, we show the evaluation results of our Drunkard's Odometry against several relevant baselines in two non-rigid datasets: our synthetic Drunkard's and the real Hamlyn data~\cite{mountney2010three,stoyanov2005soft,stoyanov2010real,pratt2010dynamic}.

\begin{table*}[!b]    
    \centering
    \scriptsize
    \setlength{\tabcolsep}{0.6pt}
    \newcommand{\groupsep}{\hskip 6pt}
    \begin{tabular}{cll cccc@{\groupsep} cccc@{\groupsep} cccc@{\groupsep} cccc}
        \toprule
        \multicolumn{3}{c}{} & \multicolumn{4}{c}{\textbf{Level 0}} & \multicolumn{4}{c}{\textbf{Level 1}} & \multicolumn{4}{c}{\textbf{Level 2}} & \multicolumn{4}{c}{\textbf{Level 3}} \\        
        \cmidrule(lr){4-7} \cmidrule(lr){8-11} \cmidrule(lr){12-15} \cmidrule(lr){16-19} 
         &  &  & frames & RPE & RPE & ATE & frames & RPE & RPE & ATE & frames & RPE & RPE & ATE & frames & RPE & RPE & ATE \\
        \multicolumn{1}{c}{\textbf{Scene}} & \multicolumn{1}{l}{\textbf{Method}} & \multirow{-2}{*}{\textbf{\makecell{Align-\\ment}}} & [\%]$\uparrow$ & [cm]$\downarrow$ & [º]$\downarrow$ & [m]$\downarrow$ & [\%]$\uparrow$ & [cm]$\downarrow$ & [º]$\downarrow$ & [m]$\downarrow$ & [\%]$\uparrow$ & [cm]$\downarrow$ & [º]$\downarrow$ & [m]$\downarrow$ & [\%]$\uparrow$ & [cm]$\downarrow$ & [º]$\downarrow$ & [m]$\downarrow$ \\            
        \midrule             
         & COLMAP~\cite{schoenberger2016sfm}  &  $\ensuremath{\mathrm{Sim}(3)}$ & 42  &   \nd 0.36  &    \fs 0.10  &   \fs 0.11  &   42  &   \nd 0.85  &   \nd 0.18  &   \fs 0.12  &   42  &   \nd 1.89  &   \nd 0.35  &   \fs 0.81  &   25  &   3.64  &   \nd 0.51  &   \nd 1.14   \\ 
         & DROID-SLAM~\cite{teed2021droid}  &                                            $\ensuremath{\mathrm{SE}(3)}$ & 100  &   \rd 0.77  &  \nd 0.28  &  2.38  &   100  &  \rd 1.41  &  \rd 0.42  &  \rd 1.79  &   100  &   2.53  &  \rd  0.69  &   \nd 1.26  &   100  &  \rd 3.58  &  \rd 1.01  &   \fs 1.00   \\ 
         & EDaM~\cite{recasens2021endo}   &    $\ensuremath{\mathrm{Sim}(3)}$ & 100  &   1.83  &  \rd 1.21  &  \rd 1.49  &   100  &   1.82  &   1.37  &   1.83  &   100  &  \rd  2.05  &   1.72  &   1.95 &   100  &   \nd 2.66  &   2.27  &   2.01  \\ 
         \multirow{-4}{*}{0}& Drunkard’s Odometry  &   $\ensuremath{\mathrm{SE}(3)}$ & 100  &   \fs 0.34  &   \fs 0.10  &   \nd 0.67  &   100  &   \fs 0.59  &   \fs 0.16  &   \nd 1.08  &  100  &   \fs 1.14  &   \fs 0.28  &  \rd 1.35 &   100  &   \fs 1.82  &   \fs 0.48  &  \rd 1.74  \\          
        \noalign{\vskip 1mm} 
        \hdashline 
        \noalign{\vskip 1mm}      
         & COLMAP~\cite{schoenberger2016sfm}  &   $\ensuremath{\mathrm{Sim}(3)}$ & 32  &   \fs 0.38  &   \fs 0.083  &   \fs 0.10  &   32  &   \nd 1.2  &   \fs 0.15  &   \fs 0.24  &   32  &   \nd 3.12  &   \nd 0.30  &   \fs 0.90  &   23  &  \rd 8.59  &   \nd 0.71  &   \fs 2.31   \\ 
         & DROID-SLAM~\cite{teed2021droid}   &  $\ensuremath{\mathrm{SE}(3)}$ &  0  &   -  &   -  &  -  &   0  &   -  &   -  &   -  &   0  &   -  &   -  &   -  &   0  &   -  &   -  &   -   \\ 
         & EDaM~\cite{recasens2021endo}   &  $\ensuremath{\mathrm{Sim}(3)}$ & 100  &  \rd 5.50  &  \rd 2.16  &  \rd 4.85  &  100  &  \rd 5.27  &  \rd 2.27  &  \rd 4.81  &   100  &   \rd 5.39  &  \rd 2.56  &  \rd 4.79 &   100  &   \nd 5.88  & \rd 2.96  &  \rd 4.90  \\ 
         \multirow{-4}{*}{4}& Drunkard’s Odometry  &     $\ensuremath{\mathrm{SE}(3)}$ & 100  &   \nd 0.60  &   \nd 0.14  &   \nd 1.21  &   100  &   \fs 0.83  &   \nd 0.18  &   \nd 1.39  &   100  &   \fs 1.43  &   \fs 0.28  &   \nd 2.46 &   100  &   \fs 2.26  &   \fs 0.46  &   \nd 4.66  \\ 
        \noalign{\vskip 1mm} 
        \hdashline 
        \noalign{\vskip 1mm}      
         & COLMAP~\cite{schoenberger2016sfm}  &   $\ensuremath{\mathrm{Sim}(3)}$ & 100  &   \fs 0.40  &   \fs 0.08  &   \fs 0.20  &   80  &   \nd 1.12  &   \fs 0.16  &   \fs 0.53  &   100  &   \rd 3.58  &   \nd 0.356  &   \nd 1.38  &   31  &   4.95  &   \fs 0.46  &   \fs 2.45   \\ 
         & DROID-SLAM~\cite{teed2021droid} &   $\ensuremath{\mathrm{SE}(3)}$ & 100  &   \rd 0.56  &   \rd 0.21  &  \rd 1.25  &   100  &   \rd 1.52  &  \rd 0.39  &  \rd 1.56  &   100  &   \nd 3.16  & \rd 0.67  &  \rd 2.43  &   100  &  \rd 4.69  &  \rd 1.02  & \nd 2.70   \\ 
         & EDaM~\cite{recasens2021endo}   &    $\ensuremath{\mathrm{Sim}(3)}$ & 100  &   3.05  &   1.98  &   2.82  &   100  &   3.13  &   2.11  &   2.73  &   100  &   3.57  &   2.46  &   2.99 &   100  &   \nd 4.12  &   2.98  &   \rd 2.86  \\ 
         \multirow{-4}{*}{5}& Drunkard’s Odometry  &   $\ensuremath{\mathrm{SE}(3)}$ & 100  &   \nd 0.45  &   \nd 0.13  &   \nd 0.47  &   100  &   \fs 0.74  &   \nd 0.18  &   \nd 0.70  &   100  &   \fs 1.44  &   \fs 0.29  &   \fs 1.24 &   100  &   \fs 2.40  &   \nd 0.49  &   \fs 2.45  \\
        \bottomrule
    \end{tabular}%
    \vspace{-6pt}
    \caption{\textbf{Trajectory errors for Drunkard's test scenes for all difficulty levels.}
    Note that COLMAP is an offline method and is only shown for reference. Our odometry method mostly outperforms the compared online odometry methods.
    Best results are highlighted as \colorbox{colorFst}{\bf first}, \colorbox{colorSnd}{second}, and \colorbox{colorTrd}{third}.}
\label{tab:table_drunk_eval}
\end{table*}

\boldparagraph{Drunkard's Setup.}
We trained our Drunkard's Odometry in all scenes of the Drunkard's Dataset except of the test ones, with around $90-10\%$ ratio for training and testing, respectively, of the difficulty level 1 with an input resolution of $320 \times 320$, batch size of $12$, learning rate of $10^{-4}$, Adam optimizer~\cite{loshchilov2017decoupled}, weight decay of $10^{-5}$, $12$ iterations of the iterative block during training and test, hyperparameters $w_1 = 0.2$, $w_2 = 100$, $w_3 = 200$ and $w_4 = 6$, and during $10$ epochs ($\sim3$ days) on a single RTX Nvidia Titan. An ablation study available in the supplementary material was performed to obtain the hyperparameter values. Everything was trained from scratch except for the pose network encoder which was pretrained on ImageNet~\cite{deng2009imagenet}.

\boldparagraph{Drunkard's Benchmark.}
We put to fight our Drunkard's Odometry in the Drunkard's Dataset benchmark against the gold-standard SfM pipeline COLMAP~\cite{schoenberger2016sfm, schoenberger2016mvs}, that searches for matches across all images in an offline manner using only the RGB channels of our images; the robust and accurate DROID-SLAM~\cite{teed2021droid} that uses a combination of local-online and global-offline bundle adjustment refinements trained in also in virtual environments with optical flow and evaluated using the same RGB-D images as Drunkard's Odometry; and the frame-to-frame tracking designed for non-rigid endoscopic scenes Endo-Depth-and-Motion (EDaM)~\cite{recasens2021endo} that here uses the RGB plus the single-view estimated depth maps using a network trained self-supervisely on monocular images of KITTI~\cite{geiger2013vision}. The non-rigid SfM SD-DefSLAM~\cite{gomez2021sd} was tested but fails in the beginning of the sequences. Complex non-rigid optimization methods such as this one are unstable and tend to fail in difficult sequences with complicated 3D surfaces and abrupt camera trajectories like the ones in our Drunkard's Dataset.

We chose scenes 0, 4 and 5 as the test ones with $3816$, $4545$ and $1655$ frames respectively. The rest of the scenes were used for training. The trajectories estimated by COLMAP and EDaM are compared against the ground truth after $\ensuremath{\mathrm{Sim}(3)}$ alignment, as they are up-to-scale. DROID-SLAM and our Drunkard's are aligned to the ground truth with a $\ensuremath{\mathrm{SE}(3)}$ transformation, as the RGB-D input allows them to estimate the real scale. The reported metrics are: Relative Position Error (RPE) for translation and rotation, that measures the local accuracy of the estimated trajectory against the ground truth between consecutive frames, and the Absolute Trajectory Error (ATE) for translation that computes the global consistency between both trajectories (see \cite{prokhorov2019measuring} for details).

For each sequence, the percentage of registered frames over the total is shown, a metric in which COLMAP shows poor performance. As a consequence, its trajectory metrics are influenced positively as it excludes frames that are challenging to track and probably would have increased the error. This is beneficial in particular for the ATE, as it takes into account the global consistency rather than frame-to-frame errors like the RPE, and happens earlier in higher deforming scenes. Also note that DROID-SLAM is very GPU-memory demanding, in part because of the final global bundle adjustment and it is not able work with long sequences like scene 4. However, our Drunkard's Odometry and EDaM are more robust, partly due to tracking only between adjacent frames.

Table~\ref{tab:table_drunk_eval} shows our results. Note that our Drunkard's Odometry practically always outperforms DROID-SLAM and EDaM in RPE and ATE, also in rigid scenes, even if our model is trained exclusively in non-rigid scenes and does not use loop closure or full bundle adjustment. 
The gap is significantly larger at higher deformation levels, for which the Drunkard's Odometry errors increase much less. This demonstrates that our method is able to generalize at predicting surface deformations. Only COLMAP is able to outperform our Drunkard's Odometry in ATE. Note, in any case, that COLMAP has a much lower recall, as it only estimates camera motion for a substantially lower percentage of frames. If we focus on RPE, a more fair metric in this case, our Drunkard's Odometry is on par to COLMAP at lower deformation levels, and clearly outperforms it at higher ones.

\begin{figure}[!b]
  \centering
  \newcommand{\imgheight}{3.1cm}
  \subfloat[][Scene 1 frame]{
  \includegraphics[height=\imgheight]{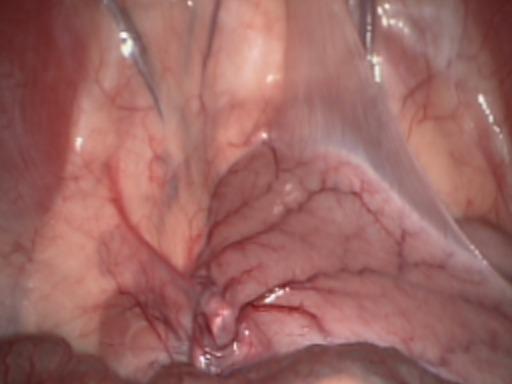}
    \label{subfig:image1}
    }
  \subfloat[][Scene 17 frame]{
    \includegraphics[height=\imgheight]{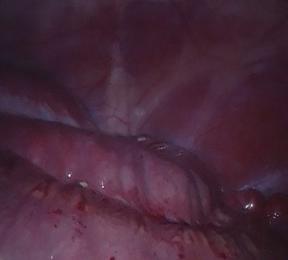}
    \label{subfig:image2}
    }
  \subfloat[][Tracking Results]{
    \scriptsize
    \setlength{\tabcolsep}{1pt}
    \begin{tabular}{cclc}
        \toprule 
        Scene & \#Frames & Method & $\mathrm{APTE}\!\downarrow$ \\        
        \midrule        
         &   &   EDaM~\cite{recasens2021endo} & \nd 0.0044  \\ 
         &   &   DROID-SLAM~\cite{teed2021droid} & \fs 0.0019  \\ 
         &   &   Drunkard’s trained w/o deform. & 0.0055  \\         
        \multirow{-4}{*}{1} & \multirow{-4}{*}{1058}   &  Drunkard’s trained w/ deform. & \rd 0.0045  \\
        \noalign{\vskip 1mm} 
        \hdashline
        \noalign{\vskip 1mm} 
        &   &   EDaM~\cite{recasens2021endo} & \nd 0.0236  \\ 
        &   &   DROID-SLAM~\cite{teed2021droid} & 0.0338  \\ 
        &   &   Drunkard’s trained w/o deform. & \rd 0.0309  \\
        \multirow{-4}{*}{17} & \multirow{-4}{*}{1306}   &   Drunkard’s trained w/ deform. & \fs 0.0214 \\
        \bottomrule
    \end{tabular}%
    \label{subfig:tableHamlyn}
    }
    \vspace{-8pt}
  \caption{\textbf{Sample frames and $\mathrm{APTE}$ results for the Hamlyn test videos.}}
  \label{fig:hamlyn}
\end{figure}

\boldparagraph{Validation in real endoscopies.}
We used the Hamlyn dataset~\cite{stoyanov2005soft}, that contains intracorporeal endoscopic RGB scenes with weak textures, deformations and reflections. Specifically, we chose scenes 1 and 17 (see Figure~\ref{fig:hamlyn}), which are significant exploratory ones. Most of Hamlyn's videos have very small camera motions, being of no interest for benchmarking odometry methods. We slightly cropped the images to remove black pixels at the borders. Depth data was taken from the public tracking test data of EDaM~\cite{recasens2021endo} which was estimated by a single-view dense depth network trained in a self-supervised manner in all Hamlyn scenes except for the test ones. Note that this depth does not have the same quality as the real ground truth one from the Drunkard's Dataset. 

Since the Hamlyn dataset lacks ground truth camera poses we propose a novel ground truth-free trajectory metric to measure the quality of the estimated odometries. 
The key idea is to generate loopy videos by duplicating and reversing a given image sequence and concatenating it to the end of the original sequence.
The generated "palindrome video" is twice as long as the original sequence and any tracking method should ideally loop back to its starting position. We then simply measure the discrepancy between start and end pose and accumulate it over various loop lengths.
We denote the new metric as the Absolute Palindrome Trajectory Error ($\mathrm{APTE}$) and define it as
\begin{equation}
   \mathrm{APTE} =  \frac{\sum_{k=1}^{N} \mathrm{APTE}_k}{N} 
   \qquad \quad \text{with} \quad
   \mathrm{APTE}_k = \left\lVert \left( \prod_{j=1}^{k} {\mathbf{T}_c}_j^{back} \right) \left(\prod_{j=1}^{k} {\mathbf{T}_c}_j\right) \right\rVert_2 \enspace,
   \label{eq:APTE}
\end{equation}
where $N$ is the total number of frames and $\mathrm{APTE}_k$ the translation Root Mean Squared Error (RMSE) of between the first and last pose of the sub-trajectory loop $k$. This last pose of the loop $k$ is the result of sequentially applying to the starting camera pose --the identity $\mathcal{I} \in \ensuremath{\mathrm{SE}(3)}$-- the first $k$ estimated relative camera poses ${\mathbf{T}_c}$ for the scene plus the first $k$ estimated relative camera poses for the same scene but ran backwards ${\mathbf{T}_c}^{back}$.
Then, the $\mathrm{APTE}_k$ is the module of the translation of this last estimated pose.
We note that the $\mathrm{APTE}$ error is trivially minimized if all transformations are zero, but this case is easy to check.

Table~\ref{subfig:tableHamlyn} shows our $\mathrm{APTE}$ metric in the Hamlyn test scenes for the baselines that are able to track the whole scene (EDaM and DROID-SLAM) and our Drunkard's Odometry trained only in rigid scenes of the Drunkard's Dataset, i.e.\ scenes from difficulty level 0, trained in deformable scenes, specifically from level 1. Before computing the $\mathrm{APTE}$, all the trajectories are scaled with the one given by a $\mathrm{Sim}(3)$ alignment, having as reference the trajectory estimated by EDaM, since it showed a good qualitative tracking performance in Endo-Depth-and-Motion~\cite{recasens2021endo}. 

\begin{figure}
\begin{subfigure}{.49\textwidth}
  \centering
  \includegraphics[width=1.\linewidth]{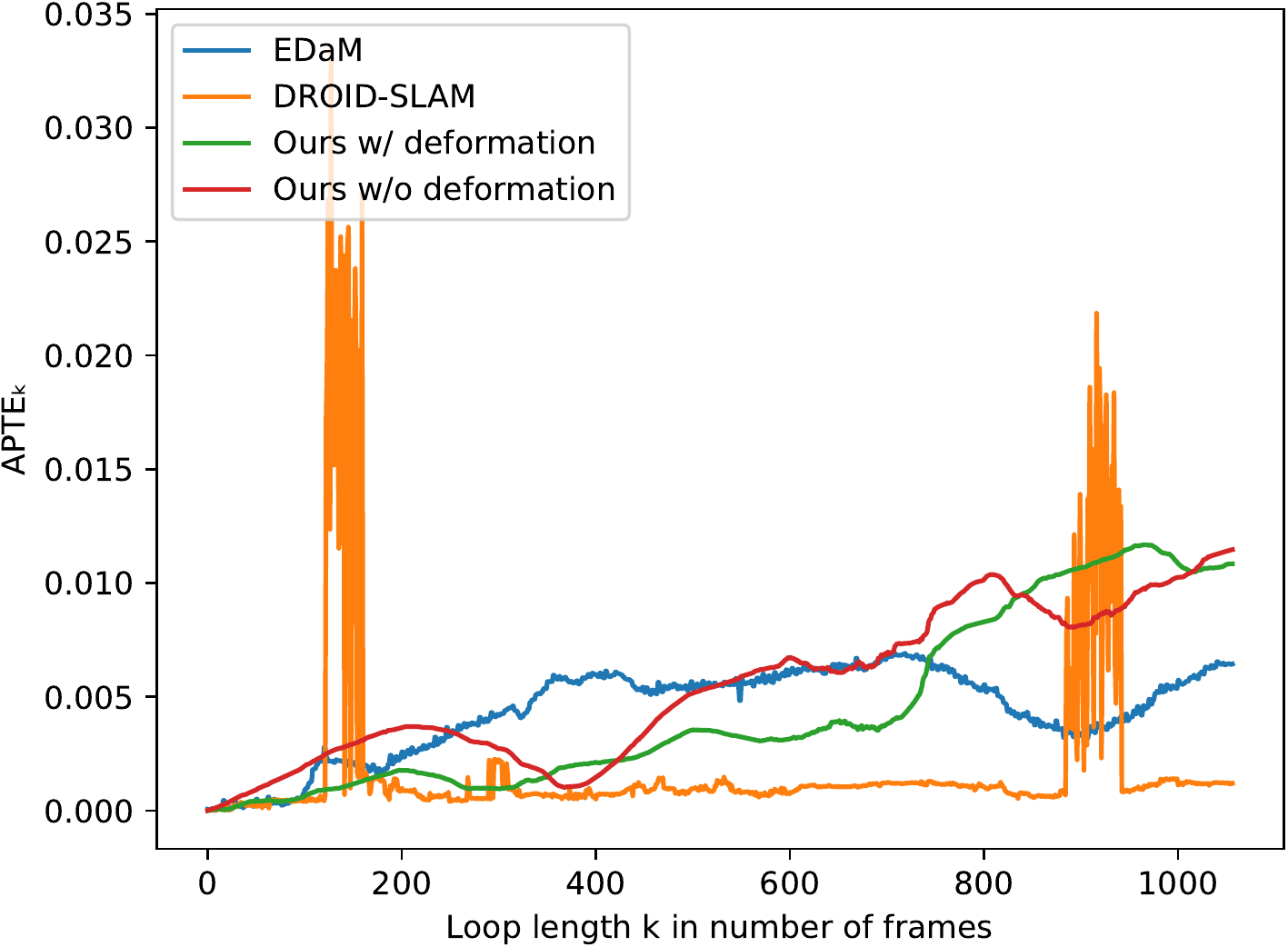}  \\[-6pt]
  \caption{\footnotesize $\mathrm{APTE}_k$ plots in Hamlyn's scene 1.
  }
    \label{fig:apte_1}

\end{subfigure}
\begin{subfigure}{.49\textwidth}
  \centering
  \includegraphics[width=1.\linewidth]{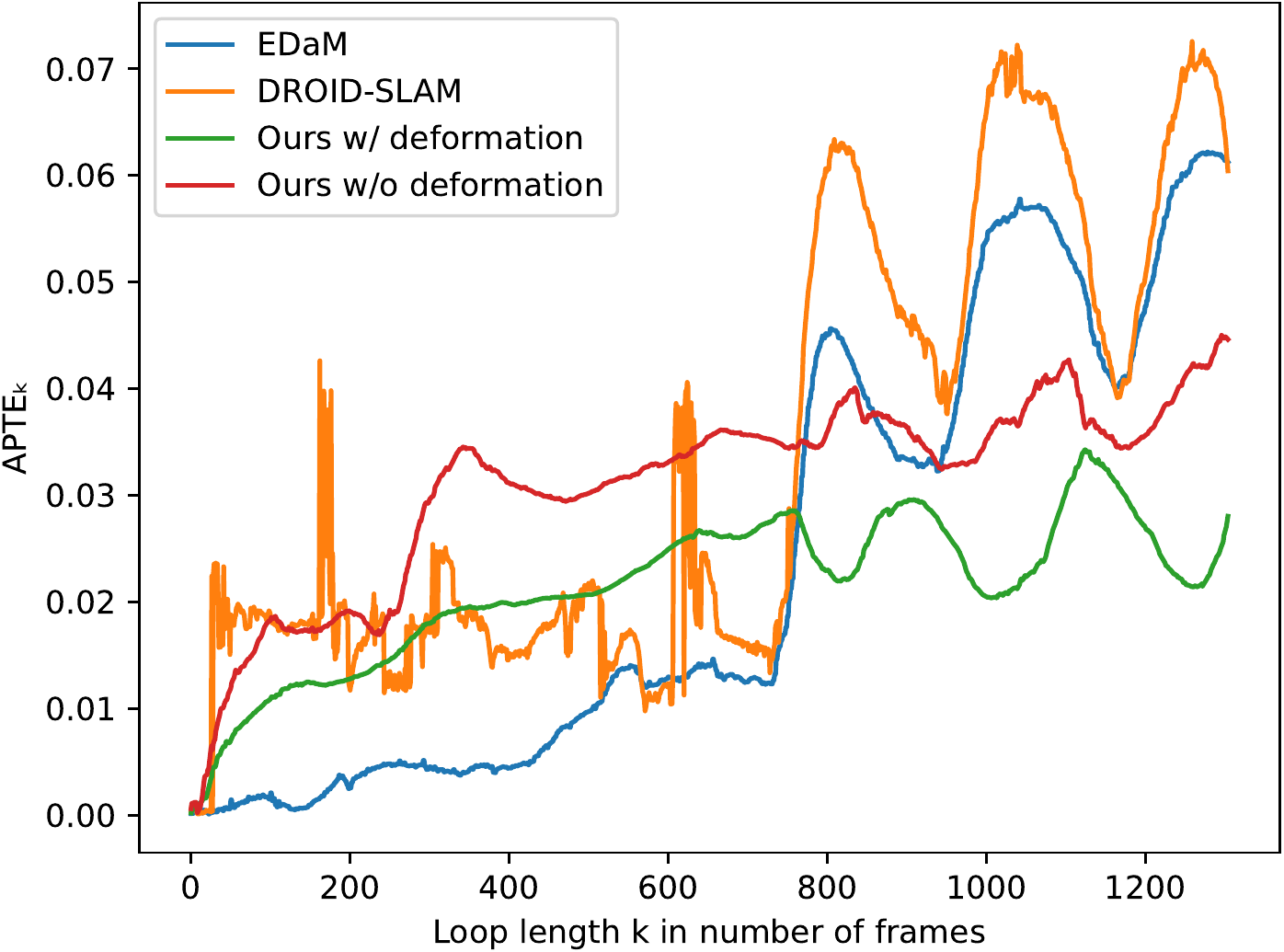}  \\[-6pt]
  \caption{\footnotesize $\mathrm{APTE}_k$ plots in Hamlyn's scene 17.}
    \label{fig:apte2}
\end{subfigure}
\vspace{-6pt}
\caption{\textbf{Camera tracking performance on the Hamlyn dataset.} $\mathrm{APTE}_k$ values along the $N$ different $k$-frames-length-loops in Hamlyn's scenes 1 (a) and 17 (b) for EDaM~\cite{recasens2021endo}, DROID-SLAM~\cite{teed2021droid} and Drunkard's Odometry (ours) with and without having been trained in deformable scenes of the Drunkard's Dataset (i.e.\ trained in level 1 and 0, respectively). The training on deforming scene substantially improves the performance of our method.}
\label{fig:apte_plots}
\end{figure}

Figure~\ref{fig:apte_plots} shows how $\mathrm{APTE}_k$ values evolve as the length of the loops grows. Notice that a longer loop does not always mean higher $\mathrm{APTE}_k$. This is because there can be some specific problematic frames with a bad pose estimation that induce a drift accumulation in the following ones getting worse $\mathrm{APTE}_k$ measures. However, this can be reversed if the tracking recovers partially during the way back in the loop. This is especially remarkable for DROID-SLAM in scene 1. It has some very unstable pose estimations, but is able to relocate later. This is probably thanks to its global bundle adjustments that readjusts all poses in the trajectory together and that works great in a scene were the same place is revisited constantly. The single $\mathrm{APTE}$ values shown in the main paper are the average of all the $\mathrm{APTE}_k$ measures.

The first conclusion we can extract is that our model performs better if it is trained in non-rigid scenes rather than rigid. Again, this shows its capacity to learn deformation patterns and retrieve a more stable camera pose under real-world non-rigid challenging scenes, and even trained in a different domain. In scene 1 the same area is revisited several times, i.e.\ there are many near loop closures always around the same place, which benefits DROID-SLAM that applies local and global bundle adjustment, hence avoiding drift. However, if there are no recurrent near loop closures and the camera is moving sharply as in scene 17, DROID-SLAM loses its advantage and EDaM and, specifically, Drunkard's Odometry outperforms it.
SD-DefSLAM~\cite{gomez2021sd} was also tested here and breaks after a few frames, far from the full lenght of the trajectory, even having originally been built and tested in Hamlyn. In consequence, we could not compute the $\mathrm{APTE}$ since we need the full trajectory ran forwards and backwards.

%
\boldparagraph{Limitations.}
\label{sec:limitations}
Our Drunkard's Dataset's most clear limitation is that it is synthetic. However, as we argued, we believe that the difficulties for acquiring high-quality data with ground truth annotations in the target application domains motivates their use. Notice that is not possible to record true ground truth optical flow in real-world sequences, as we do not have access to the exact pixel motion information. It is precisely this optical flow availability which unlocks the use of powerful flow-based models trained in synthetic deformable data which generalize well to real non-rigid scenes. Despite the real indoor images of the Drunkard's Dataset may resemble real deformable scenes in texture and shape, such as bouncy castles, moving fabrics or canvas, they are far from the medical application environment. Still, we think that generating medical data with realistic deformations, fluids or textures in large scale is out of reach. As a proof, such data does not exist yet and motivates the use of the Drunkard's data.
Our Drunkard's Odometry has all the limitations inherent to a frame-to-frame tracking method. Drift accumulates very quickly, and even if our sequences are loopy we do not either detect loop closures or correct our trajectories based on them. However, the SLAM literature shows that SLAM methods (e.g., \cite{gao2018ldso}) can be build on top of odometry ones (e.g., \cite{engel2017direct}).

\section{Conclusions}  \label{sec:conclusions}
%
Estimating camera motion in deformable scenes is a challenging research problem relatively under-explored in the literature, and for which a lack of clear benchmarks slows down research progress.
In this work, we created the Drunkard's Dataset, a large-scale simulated dataset with perfect ground truth and a wide variety of scenes and deformation levels to train and validate deep neural models.
In addition, we propose the Drunkard's Odometry method for deformable scenes to validate our dataset.
The method minimizes a scene flow loss, but as its main contribution, intrinsically decomposes the estimated twist flow into two components: The majority of motion is aimed to be explained by a rigid-body camera motion, and all remaining motion is explained by scene deformations.
In contrast to most existing works our method does not require a static scene part for estimating a reference coordinate frame which is crucial in fully deforming scenarios like endoscopy.
To also assess odometry estimates in the absence of ground truth data, we further define a novel ground-truth-free metric for trajectory evaluation that measures the cyclic consistency of a tracking algorithm.
Both the dataset and source code for our baseline method will be released upon acceptance.
Our experimental results validate our dataset, illustrates its challenges, and also shows that our Drunkard's Odometry is able to outperform relevant baselines.

{\small
\boldparagraph{Acknowledgements.}
This work was supported by the EU Comission (EU-H2020 EndoMapper GA 863146), the Spanish Government (PID2021-127685NB-I00 and TED2021-131150BI00), the Aragon Government (DGA-T45 17R/FSE), and a research grant from FIFA.
}

\balance

{\small
\bibliographystyle{ieeetr}
\bibliography{egbib}

\begin{thebibliography}{10}

\bibitem{badias2021morph}
A.~Badias, I.~Alfaro, D.~Gonzalez, F.~Chinesta, and E.~Cueto, ``Morph-dslam:
  Model order reduction for physics-based deformable slam,'' {\em IEEE
  Transactions on Pattern Analysis and Machine Intelligence}, vol.~44, no.~11,
  pp.~7764--7777, 2021.

\bibitem{li2021topologically}
T.~Li, S.~Liu, T.~Bolkart, J.~Liu, H.~Li, and Y.~Zhao, ``Topologically
  consistent multi-view face inference using volumetric sampling,'' in {\em
  Proceedings of the IEEE/CVF International Conference on Computer Vision},
  pp.~3824--3834, 2021.

\bibitem{qian2020html}
N.~Qian, J.~Wang, F.~Mueller, F.~Bernard, V.~Golyanik, and C.~Theobalt, ``Html:
  A parametric hand texture model for 3d hand reconstruction and
  personalization,'' in {\em Computer Vision--ECCV 2020: 16th European
  Conference, Glasgow, UK, August 23--28, 2020, Proceedings, Part XI 16},
  pp.~54--71, Springer, 2020.

\bibitem{wang2021deep}
J.~Wang, S.~Tan, X.~Zhen, S.~Xu, F.~Zheng, Z.~He, and L.~Shao, ``Deep 3d human
  pose estimation: A review,'' {\em Computer Vision and Image Understanding},
  vol.~210, p.~103225, 2021.

\bibitem{li2021coarse}
C.~Li and G.~H. Lee, ``Coarse-to-fine animal pose and shape estimation,'' {\em
  Advances in Neural Information Processing Systems}, vol.~34,
  pp.~11757--11768, 2021.

\bibitem{su2020mulaycap}
Z.~Su, W.~Wan, T.~Yu, L.~Liu, L.~Fang, W.~Wang, and Y.~Liu, ``Mulaycap:
  Multi-layer human performance capture using a monocular video camera,'' {\em
  IEEE Transactions on Visualization and Computer Graphics}, vol.~28, no.~4,
  pp.~1862--1879, 2020.

\bibitem{lamarca2020defslam}
J.~Lamarca, S.~Parashar, A.~Bartoli, and J.~Montiel, ``Defslam: Tracking and
  mapping of deforming scenes from monocular sequences,'' {\em IEEE
  Transactions on robotics}, vol.~37, no.~1, pp.~291--303, 2020.

\bibitem{sengupta2021colonoscopic}
A.~Sengupta and A.~Bartoli, ``Colonoscopic 3d reconstruction by tubular
  non-rigid structure-from-motion,'' {\em International Journal of Computer
  Assisted Radiology and Surgery}, vol.~16, no.~7, pp.~1237--1241, 2021.

\bibitem{recasens2021endo}
D.~Recasens, J.~Lamarca, J.~M. F{\'a}cil, J.~Montiel, and J.~Civera,
  ``Endo-depth-and-motion: reconstruction and tracking in endoscopic videos
  using depth networks and photometric constraints,'' {\em IEEE Robotics and
  Automation Letters}, vol.~6, no.~4, pp.~7225--7232, 2021.

\bibitem{shao2022self}
S.~Shao, Z.~Pei, W.~Chen, W.~Zhu, X.~Wu, D.~Sun, and B.~Zhang,
  ``Self-supervised monocular depth and ego-motion estimation in endoscopy:
  Appearance flow to the rescue,'' {\em Medical image analysis}, vol.~77,
  p.~102338, 2022.

\bibitem{rabaud2008re}
V.~Rabaud and S.~Belongie, ``Re-thinking non-rigid structure from motion,'' in
  {\em 2008 IEEE Conference on Computer Vision and Pattern Recognition},
  pp.~1--8, IEEE, 2008.

\bibitem{gotardo2011non}
P.~F. Gotardo and A.~M. Martinez, ``Non-rigid structure from motion with
  complementary rank-3 spaces,'' in {\em CVPR 2011}, pp.~3065--3072, IEEE,
  2011.

\bibitem{lee2013procrustean}
M.~Lee, J.~Cho, C.-H. Choi, and S.~Oh, ``Procrustean normal distribution for
  non-rigid structure from motion,'' in {\em Proceedings of the IEEE Conference
  on computer vision and pattern recognition}, pp.~1280--1287, 2013.

\bibitem{garg2013dense}
R.~Garg, A.~Roussos, and L.~Agapito, ``Dense variational reconstruction of
  non-rigid surfaces from monocular video,'' in {\em Proceedings of the IEEE
  Conference on computer vision and pattern recognition}, pp.~1272--1279, 2013.

\bibitem{dai2014simple}
Y.~Dai, H.~Li, and M.~He, ``A simple prior-free method for non-rigid
  structure-from-motion factorization,'' {\em International Journal of Computer
  Vision}, vol.~107, pp.~101--122, 2014.

\bibitem{agudo2014good}
A.~Agudo, L.~Agapito, B.~Calvo, and J.~M. Montiel, ``Good vibrations: A modal
  analysis approach for sequential non-rigid structure from motion,'' in {\em
  Proceedings of the IEEE Conference on computer vision and pattern
  recognition}, pp.~1558--1565, 2014.

\bibitem{kong2019deep}
C.~Kong and S.~Lucey, ``Deep non-rigid structure from motion,'' in {\em
  Proceedings of the IEEE/CVF International Conference on Computer Vision},
  pp.~1558--1567, 2019.

\bibitem{bozic2020deepdeform}
A.~Bozic, M.~Zollhofer, C.~Theobalt, and M.~Nie{\ss}ner, ``Deepdeform: Learning
  non-rigid rgb-d reconstruction with semi-supervised data,'' in {\em
  Proceedings of the IEEE/CVF Conference on Computer Vision and Pattern
  Recognition}, pp.~7002--7012, 2020.

\bibitem{yang2021lasr}
G.~Yang, D.~Sun, V.~Jampani, D.~Vlasic, F.~Cole, H.~Chang, D.~Ramanan, W.~T.
  Freeman, and C.~Liu, ``Lasr: Learning articulated shape reconstruction from a
  monocular video,'' in {\em Proceedings of the IEEE/CVF Conference on Computer
  Vision and Pattern Recognition}, pp.~15980--15989, 2021.

\bibitem{tretschk2021non}
E.~Tretschk, A.~Tewari, V.~Golyanik, M.~Zollh{\"o}fer, C.~Lassner, and
  C.~Theobalt, ``Non-rigid neural radiance fields: Reconstruction and novel
  view synthesis of a dynamic scene from monocular video,'' in {\em Proceedings
  of the IEEE/CVF International Conference on Computer Vision},
  pp.~12959--12970, 2021.

\bibitem{pumarola2021d}
A.~Pumarola, E.~Corona, G.~Pons-Moll, and F.~Moreno-Noguer, ``D-nerf: Neural
  radiance fields for dynamic scenes,'' in {\em Proceedings of the IEEE/CVF
  Conference on Computer Vision and Pattern Recognition}, pp.~10318--10327,
  2021.

\bibitem{potje2021extracting}
G.~Potje, R.~Martins, F.~Chamone, and E.~Nascimento, ``Extracting
  deformation-aware local features by learning to deform,'' {\em Advances in
  Neural Information Processing Systems}, vol.~34, pp.~10759--10771, 2021.

\bibitem{parashar2021robust}
S.~Parashar, D.~Pizarro, and A.~Bartoli, ``Robust isometric non-rigid
  structure-from-motion,'' {\em IEEE Transactions on Pattern Analysis and
  Machine Intelligence}, vol.~44, no.~10, pp.~6409--6423, 2021.

\bibitem{weng2022humannerf}
C.-Y. Weng, B.~Curless, P.~P. Srinivasan, J.~T. Barron, and
  I.~Kemelmacher-Shlizerman, ``Humannerf: Free-viewpoint rendering of moving
  people from monocular video,'' in {\em Proceedings of the IEEE/CVF Conference
  on Computer Vision and Pattern Recognition}, pp.~16210--16220, 2022.

\bibitem{cadena2016past}
C.~Cadena, L.~Carlone, H.~Carrillo, Y.~Latif, D.~Scaramuzza, J.~Neira, I.~Reid,
  and J.~J. Leonard, ``Past, present, and future of simultaneous localization
  and mapping: Toward the robust-perception age,'' {\em IEEE Transactions on
  robotics}, vol.~32, no.~6, pp.~1309--1332, 2016.

\bibitem{gomez2021sd}
J.~J. G{\'o}mez-Rodr{\'\i}guez, J.~Lamarca, J.~Morlana, J.~D. Tard{\'o}s, and
  J.~M. Montiel, ``Sd-defslam: Semi-direct monocular slam for deformable and
  intracorporeal scenes,'' in {\em 2021 IEEE International Conference on
  Robotics and Automation (ICRA)}, pp.~5170--5177, IEEE, 2021.

\bibitem{ramakrishnan2021habitat}
S.~K. Ramakrishnan, A.~Gokaslan, E.~Wijmans, O.~Maksymets, A.~Clegg, J.~Turner,
  E.~Undersander, W.~Galuba, A.~Westbury, A.~X. Chang, {\em et~al.},
  ``Habitat-matterport 3d dataset (hm3d): 1000 large-scale 3d environments for
  embodied ai,'' {\em arXiv preprint arXiv:2109.08238}, 2021.

\bibitem{butler2012naturalistic}
D.~J. Butler, J.~Wulff, G.~B. Stanley, and M.~J. Black, ``A naturalistic open
  source movie for optical flow evaluation,'' in {\em Computer Vision--ECCV
  2012: 12th European Conference on Computer Vision, Florence, Italy, October
  7-13, 2012, Proceedings, Part VI 12}, pp.~611--625, Springer, 2012.

\bibitem{mayer2016large}
N.~Mayer, E.~Ilg, P.~Hausser, P.~Fischer, D.~Cremers, A.~Dosovitskiy, and
  T.~Brox, ``A large dataset to train convolutional networks for disparity,
  optical flow, and scene flow estimation,'' in {\em Proceedings of the IEEE
  conference on computer vision and pattern recognition}, pp.~4040--4048, 2016.

\bibitem{stoyanov2005soft}
D.~Stoyanov, G.~P. Mylonas, F.~Deligianni, A.~Darzi, and G.-Z. Yang,
  ``Soft-tissue motion tracking and structure estimation for robotic assisted
  mis procedures,'' in {\em MICCAI (2)}, pp.~139--146, 2005.

\bibitem{stoyanov2012stereoscopic}
D.~Stoyanov, ``Stereoscopic scene flow for robotic assisted minimally invasive
  surgery,'' in {\em Medical Image Computing and Computer-Assisted
  Intervention--MICCAI 2012: 15th International Conference, Nice, France,
  October 1-5, 2012, Proceedings, Part I 15}, pp.~479--486, Springer, 2012.

\bibitem{jensen2021benchmark}
S.~H.~N. Jensen, M.~E.~B. Doest, H.~Aan{\ae}s, and A.~Del~Bue, ``A benchmark
  and evaluation of non-rigid structure from motion,'' {\em International
  Journal of Computer Vision}, vol.~129, no.~4, pp.~882--899, 2021.

\bibitem{azagra2022endomapper}
P.~Azagra, C.~Sostres, {\'A}.~Ferrandez, L.~Riazuelo, C.~Tomasini, O.~L.
  Barbed, J.~Morlana, D.~Recasens, V.~M. Batlle, J.~J.
  G{\'o}mez-Rodr{\'\i}guez, {\em et~al.}, ``Endomapper dataset of complete
  calibrated endoscopy procedures,'' {\em arXiv preprint arXiv:2204.14240},
  2022.

\bibitem{teed2021droid}
Z.~Teed and J.~Deng, ``Droid-slam: Deep visual slam for monocular, stereo, and
  rgb-d cameras,'' {\em Advances in Neural Information Processing Systems},
  vol.~34, 2021.

\bibitem{teed2021raft}
Z.~Teed and J.~Deng, ``Raft-3d: Scene flow using rigid-motion embeddings,'' in
  {\em Proceedings of the IEEE/CVF Conference on Computer Vision and Pattern
  Recognition}, pp.~8375--8384, 2021.

\bibitem{de2008performance}
E.~De~Aguiar, C.~Stoll, C.~Theobalt, N.~Ahmed, H.-P. Seidel, and S.~Thrun,
  ``Performance capture from sparse multi-view video,'' in {\em ACM SIGGRAPH
  2008 papers}, pp.~1--10, 2008.

\bibitem{bogo2014faust}
F.~Bogo, J.~Romero, M.~Loper, and M.~J. Black, ``Faust: Dataset and evaluation
  for 3d mesh registration,'' in {\em Proceedings of the IEEE conference on
  computer vision and pattern recognition}, pp.~3794--3801, 2014.

\bibitem{li20214dcomplete}
Y.~Li, H.~Takehara, T.~Taketomi, B.~Zheng, and M.~Nie{\ss}ner, ``4dcomplete:
  Non-rigid motion estimation beyond the observable surface,'' in {\em
  Proceedings of the IEEE/CVF International Conference on Computer Vision},
  pp.~12706--12716, 2021.

\bibitem{mountney2010three}
P.~Mountney, D.~Stoyanov, and G.-Z. Yang, ``Three-dimensional tissue
  deformation recovery and tracking,'' {\em IEEE Signal Processing Magazine},
  vol.~27, no.~4, pp.~14--24, 2010.

\bibitem{stoyanov2010real}
D.~Stoyanov, M.~V. Scarzanella, P.~Pratt, and G.-Z. Yang, ``Real-time stereo
  reconstruction in robotically assisted minimally invasive surgery,'' in {\em
  International Conference on Medical Image Computing and Computer-Assisted
  Intervention}, 2010.

\bibitem{pratt2010dynamic}
P.~Pratt, D.~Stoyanov, M.~Visentini-Scarzanella, and G.-Z. Yang, ``Dynamic
  guidance for robotic surgery using image-constrained biomechanical models,''
  in {\em International Conference on Medical Image Computing and
  Computer-Assisted Intervention}, pp.~77--85, 2010.

\bibitem{sturm2012benchmark}
J.~Sturm, N.~Engelhard, F.~Endres, W.~Burgard, and D.~Cremers, ``A benchmark
  for the evaluation of rgb-d slam systems,'' in {\em 2012 IEEE/RSJ
  international conference on intelligent robots and systems}, pp.~573--580,
  IEEE, 2012.

\bibitem{geiger2013vision}
A.~Geiger, P.~Lenz, C.~Stiller, and R.~Urtasun, ``Vision meets robotics: The
  kitti dataset,'' {\em The International Journal of Robotics Research},
  vol.~32, no.~11, pp.~1231--1237, 2013.

\bibitem{wilson2014robust}
K.~Wilson and N.~Snavely, ``Robust global translations with 1dsfm,'' in {\em
  Computer Vision--ECCV 2014: 13th European Conference, Zurich, Switzerland,
  September 6-12, 2014, Proceedings, Part III 13}, pp.~61--75, Springer, 2014.

\bibitem{burri2016euroc}
M.~Burri, J.~Nikolic, P.~Gohl, T.~Schneider, J.~Rehder, S.~Omari, M.~W.
  Achtelik, and R.~Siegwart, ``The euroc micro aerial vehicle datasets,'' {\em
  The International Journal of Robotics Research}, vol.~35, no.~10,
  pp.~1157--1163, 2016.

\bibitem{dai2017scannet}
A.~Dai, A.~X. Chang, M.~Savva, M.~Halber, T.~Funkhouser, and M.~Nie{\ss}ner,
  ``Scannet: Richly-annotated 3d reconstructions of indoor scenes,'' in {\em
  Proceedings of the IEEE conference on computer vision and pattern
  recognition}, pp.~5828--5839, 2017.

\bibitem{schubert2018tum}
D.~Schubert, T.~Goll, N.~Demmel, V.~Usenko, J.~St{\"u}ckler, and D.~Cremers,
  ``The tum vi benchmark for evaluating visual-inertial odometry,'' in {\em
  2018 IEEE/RSJ International Conference on Intelligent Robots and Systems
  (IROS)}, pp.~1680--1687, IEEE, 2018.

\bibitem{schops2019bad}
T.~Schops, T.~Sattler, and M.~Pollefeys, ``Bad slam: Bundle adjusted direct
  rgb-d slam,'' in {\em Proceedings of the IEEE/CVF Conference on Computer
  Vision and Pattern Recognition}, pp.~134--144, 2019.

\bibitem{handa2014benchmark}
A.~Handa, T.~Whelan, J.~McDonald, and A.~J. Davison, ``A benchmark for rgb-d
  visual odometry, 3d reconstruction and slam,'' in {\em 2014 IEEE
  international conference on Robotics and automation (ICRA)}, pp.~1524--1531,
  IEEE, 2014.

\bibitem{straub2019replica}
J.~Straub, T.~Whelan, L.~Ma, Y.~Chen, E.~Wijmans, S.~Green, J.~J. Engel,
  R.~Mur-Artal, C.~Ren, S.~Verma, {\em et~al.}, ``The replica dataset: A
  digital replica of indoor spaces,'' {\em arXiv preprint arXiv:1906.05797},
  2019.

\bibitem{wang2020tartanair}
W.~Wang, D.~Zhu, X.~Wang, Y.~Hu, Y.~Qiu, C.~Wang, Y.~Hu, A.~Kapoor, and
  S.~Scherer, ``Tartanair: A dataset to push the limits of visual slam,'' in
  {\em 2020 IEEE/RSJ International Conference on Intelligent Robots and Systems
  (IROS)}, pp.~4909--4916, IEEE, 2020.

\bibitem{minoda2021viode}
K.~Minoda, F.~Schilling, V.~W{\"u}est, D.~Floreano, and T.~Yairi, ``Viode: A
  simulated dataset to address the challenges of visual-inertial odometry in
  dynamic environments,'' {\em IEEE Robotics and Automation Letters}, vol.~6,
  no.~2, pp.~1343--1350, 2021.

\bibitem{baker2011database}
S.~Baker, D.~Scharstein, J.~Lewis, S.~Roth, M.~J. Black, and R.~Szeliski, ``A
  database and evaluation methodology for optical flow,'' {\em International
  journal of computer vision}, vol.~92, pp.~1--31, 2011.

\bibitem{dosovitskiy2015flownet}
A.~Dosovitskiy, P.~Fischer, E.~Ilg, P.~Hausser, C.~Hazirbas, V.~Golkov, P.~Van
  Der~Smagt, D.~Cremers, and T.~Brox, ``Flownet: Learning optical flow with
  convolutional networks,'' in {\em Proceedings of the IEEE international
  conference on computer vision}, pp.~2758--2766, 2015.

\bibitem{ilg2018occlusions}
E.~Ilg, T.~Saikia, M.~Keuper, and T.~Brox, ``Occlusions, motion and depth
  boundaries with a generic network for disparity, optical flow or scene flow
  estimation,'' in {\em Proceedings of the European conference on computer
  vision (ECCV)}, pp.~614--630, 2018.

\bibitem{mehl2023spring}
L.~Mehl, J.~Schmalfuss, A.~Jahedi, Y.~Nalivayko, and A.~Bruhn, ``Spring: A
  high-resolution high-detail dataset and benchmark for scene flow, optical
  flow and stereo,'' {\em arXiv preprint arXiv:2303.01943}, 2023.

\bibitem{carlucci2017deep}
F.~M. Carlucci, P.~Russo, and B.~Caputo, ``A deep representation for depth
  images from synthetic data,'' in {\em 2017 IEEE international conference on
  robotics and automation (ICRA)}, pp.~1362--1369, IEEE, 2017.

\bibitem{planche2017depthsynth}
B.~Planche, Z.~Wu, K.~Ma, S.~Sun, S.~Kluckner, O.~Lehmann, T.~Chen, A.~Hutter,
  S.~Zakharov, H.~Kosch, {\em et~al.}, ``Depthsynth: Real-time realistic
  synthetic data generation from cad models for 2.5 d recognition,'' in {\em
  2017 International Conference on 3D Vision (3DV)}, pp.~1--10, IEEE, 2017.

\bibitem{chang2015shapenet}
A.~X. Chang, T.~Funkhouser, L.~Guibas, P.~Hanrahan, Q.~Huang, Z.~Li,
  S.~Savarese, M.~Savva, S.~Song, H.~Su, {\em et~al.}, ``Shapenet: An
  information-rich 3d model repository,'' {\em arXiv preprint
  arXiv:1512.03012}, 2015.

\bibitem{gaidon2016virtual}
A.~Gaidon, Q.~Wang, Y.~Cabon, and E.~Vig, ``Virtual worlds as proxy for
  multi-object tracking analysis,'' in {\em Proceedings of the IEEE conference
  on computer vision and pattern recognition}, pp.~4340--4349, 2016.

\bibitem{mccormac2017scenenet}
J.~McCormac, A.~Handa, S.~Leutenegger, and A.~J. Davison, ``Scenenet rgb-d: Can
  5m synthetic images beat generic imagenet pre-training on indoor
  segmentation?,'' in {\em Proceedings of the IEEE International Conference on
  Computer Vision}, pp.~2678--2687, 2017.

\bibitem{johnson2017clevr}
J.~Johnson, B.~Hariharan, L.~Van Der~Maaten, L.~Fei-Fei, C.~Lawrence~Zitnick,
  and R.~Girshick, ``Clevr: A diagnostic dataset for compositional language and
  elementary visual reasoning,'' in {\em Proceedings of the IEEE conference on
  computer vision and pattern recognition}, pp.~2901--2910, 2017.

\bibitem{roberts2021hypersim}
M.~Roberts, J.~Ramapuram, A.~Ranjan, A.~Kumar, M.~A. Bautista, N.~Paczan,
  R.~Webb, and J.~M. Susskind, ``Hypersim: A photorealistic synthetic dataset
  for holistic indoor scene understanding,'' in {\em Proceedings of the
  IEEE/CVF international conference on computer vision}, pp.~10912--10922,
  2021.

\bibitem{nikolenko2019synthetic}
S.~I. Nikolenko, ``Synthetic data for deep learning,'' {\em arXiv preprint
  arXiv:1909.11512}, 2019.

\bibitem{song2018mis}
J.~Song, J.~Wang, L.~Zhao, S.~Huang, and G.~Dissanayake, ``Mis-slam: Real-time
  large-scale dense deformable slam system in minimal invasive surgery based on
  heterogeneous computing,'' {\em IEEE Robotics and Automation Letters},
  vol.~3, no.~4, pp.~4068--4075, 2018.

\bibitem{lamarca2022direct}
J.~Lamarca, J.~J.~G. Rodriguez, J.~D. Tardos, and J.~M. Montiel, ``Direct and
  sparse deformable tracking,'' {\em IEEE Robotics and Automation Letters},
  vol.~7, no.~4, pp.~11450--11457, 2022.

\bibitem{rodriguez2022tracking}
J.~J.~G. Rodr{\'\i}guez, J.~M. Montiel, and J.~D. Tard{\'o}s, ``Tracking
  monocular camera pose and deformation for slam inside the human body,'' in
  {\em 2022 IEEE/RSJ International Conference on Intelligent Robots and Systems
  (IROS)}, pp.~5278--5285, IEEE, 2022.

\bibitem{blender}
B.~O. Community, {\em Blender - a 3D modelling and rendering package}.
\newblock Blender Foundation, Stichting Blender Foundation, Amsterdam, 2018.

\bibitem{cho2014properties}
K.~Cho, B.~Van~Merri{\"e}nboer, D.~Bahdanau, and Y.~Bengio, ``On the properties
  of neural machine translation: Encoder-decoder approaches,'' {\em arXiv
  preprint arXiv:1409.1259}, 2014.

\bibitem{schoenberger2016sfm}
J.~L. Sch\"{o}nberger and J.-M. Frahm, ``Structure-from-motion revisited,'' in
  {\em Conference on Computer Vision and Pattern Recognition (CVPR)}, 2016.

\bibitem{loshchilov2017decoupled}
I.~Loshchilov and F.~Hutter, ``Decoupled weight decay regularization,'' {\em
  arXiv preprint arXiv:1711.05101}, 2017.

\bibitem{deng2009imagenet}
J.~Deng, W.~Dong, R.~Socher, L.-J. Li, K.~Li, and L.~Fei-Fei, ``Imagenet: A
  large-scale hierarchical image database,'' in {\em 2009 IEEE conference on
  computer vision and pattern recognition}, pp.~248--255, Ieee, 2009.

\bibitem{schoenberger2016mvs}
J.~L. Sch\"{o}nberger, E.~Zheng, M.~Pollefeys, and J.-M. Frahm, ``Pixelwise
  view selection for unstructured multi-view stereo,'' in {\em European
  Conference on Computer Vision (ECCV)}, 2016.

\bibitem{prokhorov2019measuring}
D.~Prokhorov, D.~Zhukov, O.~Barinova, K.~Anton, and A.~Vorontsova, ``Measuring
  robustness of visual slam,'' in {\em 2019 16th International Conference on
  Machine Vision Applications (MVA)}, pp.~1--6, IEEE, 2019.

\bibitem{gao2018ldso}
X.~Gao, R.~Wang, N.~Demmel, and D.~Cremers, ``Ldso: Direct sparse odometry with
  loop closure,'' in {\em 2018 IEEE/RSJ International Conference on Intelligent
  Robots and Systems (IROS)}, pp.~2198--2204, IEEE, 2018.

\bibitem{engel2017direct}
J.~Engel, V.~Koltun, and D.~Cremers, ``Direct sparse odometry,'' {\em IEEE
  transactions on pattern analysis and machine intelligence}, vol.~40, no.~3,
  pp.~611--625, 2017.

\end{thebibliography}
}
 
\ifarXiv
    

\section*{Supplementary material}

# Supplementary Material for The Drunkard's Odometry: Estimating Camera Motion in Deforming Scenes

**David Recasens**
University of Zaragoza

**Martin R. Oswald**
ETH Zurich, University of Amsterdam

**Marc Pollefeys**
ETH Zurich, Microsoft

**Javier Civera**
University of Zaragoza

## Abstract

This supplementary document provides additional details about the dataset, the method, the evaluation and further experimental results.

## Contents

## A    Drunkard's Dataset Details

Table A.1 details the number of frames for each individual scene of the Drunkard's Dataset, and the total one. The number of frames is equal across all difficulty levels (hence the $4\times$ in the table). Aggregating all the levels, the Drunkard's Dataset has more than 400K high-resolution frames. In order to enhance its suitability with different computing setups and to reduce its overall size, the dataset is made available in two different image resolutions, $1024 \times 1024$ and $320 \times 320$ pixels. Other resolutions can be easily regenerated using the provided scripts. The Drunkard's Dataset is built by animating the 3D meshes of the Habitat-Matterport 3D dataset (1) (that is MIT licensed) and we use the same scene numbering. Scene 17 was skipped since it did not contain an appropriate 3D model to create an exploratory camera trajectory in it. The Blender files are also made public with scripts to replicate the rendering, so additionally anyone can create their custom datasets with different deformations, resolution, camera parameters or trajectory for example.

## B    Drunkard's Odometry Flow

Table B.1 shows 2D and 3D flow evaluation metrics for Drunkard's Odometry and its inspirational work RAFT-3D (2) (which is BSD 3-clause licensed) in all the four levels of the same four test scenes of the Drunkard's Dataset. As already said in the main paper, RAFT-3D exclusively predicts 3D scene flow, while we additionally estimate the camera pose. We evaluate both methods in terms of flow metrics in order to validate our implementation.

Both methods are trained from scratch during 10 epochs on the same Drunkard's Dataset scenes (all except the test ones) and with the same deformation (difficulty level 1). The base evaluations

| Scene | # Frames | Scene | # Frames | Scene | # Frames |
|---|---|---|---|---|---|
| 0 | 4 × 3.816 | 7 | 4 × 2.341 | 14 | 4 × 23.863 |
| 1 | 4 × 1.523 | 8 | 4 × 3.033 | 15 | 4 × 7.984 |
| 2 | 4 × 1.240 | 9 | 4 × 8.098 | 16 | 4 × 5.389 |
| 3 | 4 × 3.308 | 10 | 4 × 2.632 | 18 | 4 × 10.989 |
| 4 | 4 × 4.545 | 11 | 4 × 4.168 | 19 | 4 × 9.739 |
| 5 | 4 × 1.655 | 12 | 4 × 4.168 | **Total** | **4 × 104.302** |
| 6 | 4 × 1.515 | 13 | 4 × 3.296 | | |

Table A.1: **Drunkard's dataset overview.** The datasets consists of more than 400K frames distributed over 20 scenes and 4 difficulty levels. The table lists the number of frames for every scene and the total for one of the four (4×) difficulty levels.

metrics are the 2D and 3D End-Point-Error (EPE). EPE is computed as the mean Euclidean distance between the estimated and ground truth flow (**flow** and $\overline{\textbf{flow}}$, respectively) in all valid pixels of the $N$ test images

$$\text{EPE} = \frac{\sum_{i=1}^{N}\|\textbf{flow} - \overline{\textbf{flow}}\|_2}{N} \quad , \tag{1}$$

In the 2D case, **flow** is the optical flow that we denote in the main paper as $\mathbf{O}$. In the 3D case, **flow** is the 3D scene flow $\textbf{flow}_{3D}$ that comes from back-projecting every pixel $j \in \mathbf{I}_1$ with coordinates $\mathbf{u}_j$ and its corresponding in camera 2 $\mathbf{u}_{j'}$, associated by the optical flow $\mathbf{O}$, to the 3D space using their sensor depth values

$$\textbf{flow}_{3D} = \mathbf{P}_{j'} - \mathbf{P}_j = \pi^{-1}\left(\mathbf{u}_{j'}, \bar{z}_{j'}\right) - \pi^{-1}\left(\mathbf{u}_j, \bar{z}_j\right) \quad . \tag{2}$$

We exclude invalid pixels which have excessively large optical flow module values ($> 250$), infinite depths (as they define holes in the 3D mesh), as well as pixels for which correspondences fall outside the image boundaries.

The $\delta$ metrics are the percentage of the total pixels in the test scenes that have an EPE value under a certain threshold, being 1 pixel for the 2D case and 1 cm for the 3D case. In the 3D metrics, $|\Delta\text{EPE}|$ [%] stands for the absolute difference in percentage between the 3D EPE of RAFT-3D and of Drunkard's Odometry.

From EPE and $\delta$ metrics in Table B.1 we conclude that our Drunkard's Odometry, even focused on camera pose and not on flow as RAFT-3D, performs close to RAFT-3D on these metrics. As RAFT-3D estimates only flow, it is not as affected as our Drunkard's Odometry by the different levels of deformation. The stability of the $|\Delta\text{EPE}|$ [%] values for the four difficulty levels at each scene shows that the 3D flow estimation of our Drunkard's Odometry is stable. This gives evidence about the capability of our method to simultaneously estimate accurate scene flow $\mathbf{T}$ and camera poses $\mathbf{T}_c$ independently of the deformations, as the total flow $\mathbf{T}$ includes the flow caused by scene deformations $\mathbf{T}_d$ and the one caused by camera motion $\mathbf{T}_c$.

## C  Drunkard's Odometry Ablations

We tried several variations of our Drunkard's Odometry to find the best training setup, which was possible thanks to our generated dataset. Figure C.1 shows the Relative Position Error (RPE) metric for translation and rotation (without trajectory alignment as the training dataloader samples random adjacent pair of frames) in the three Drunkard's Dataset test scenes at level of difficulty 1 for different variations of the Drunkard's Odometry. All the methods were trained during 10 epochs from scratch, but varying training parameters. Specifically, we played with the learning rate (l.r., $10^{-5}$, $10^{-4}$ and $2 \times 10^{-4}$), inverting the order of the input image pair randomly (with 50% probability), and different loss hyperparameters. We also trained models deactivating the intermediate optical flow $\mathbf{O}^{k,\text{pre}}$ loss term with $w_1$, changing the influence of the pre-estimate of the camera pose $\mathbf{T}_{\text{pose}}^{\text{pre}}$ $w_4$, cancelling all supervision from optical flow and depth ground truth leaving only the pose loss, and modifying the relative weight $w_{\text{rot}}$ between translation and rotation in the pose loss (that by default is 1).

$$\|\log_{\text{SE}(3)}\left(\mathbf{T}_c^{\text{pre}}\bar{\mathbf{T}}_c^{-1}\right)\|_1 = \|\log_{\text{SE}(3)}\left(\mathbf{T}_c^{\text{pre}}\bar{\mathbf{T}}_c^{-1}\right)\|_1^{\text{tra}} + w_{\text{rot}}\|\log_{\text{SE}(3)}\left(\mathbf{T}_c^{\text{pre}}\bar{\mathbf{T}}_c^{-1}\right)\|_1^{\text{rot}} \quad . \tag{3}$$

| Scene | Level | Method | 2D Metrics | | 3D Metrics | | |
|---|---|---|---|---|---|---|---|
| | | | $\delta < 1$ px [%]↓ | EPE [px]↓ | $\delta < 1$ cm [%]↓ | EPE [mm]↓ | $|\Delta\text{EPE}|$ [%] |
| 0 | 0 | RAFT-3D | 93.1 | 0.43 | 96.1 | 2.77 | 32.7 |
| | | Drunkard's Odometry | 91.5 | 0.56 | 95.0 | 3.67 | |
| | 1 | RAFT-3D | 91.8 | 0.48 | 95.4 | 3.06 | 33.8 |
| | | Drunkard's Odometry | 89.5 | 0.63 | 94.1 | 4.11 | |
| | 2 | RAFT-3D | 87.2 | 0.69 | 93.1 | 3.92 | 34.3 |
| | | Drunkard's Odometry | 83.7 | 0.88 | 90.9 | 5.27 | |
| | 3 | RAFT-3D | 79.1 | 1.29 | 89.2 | 5.52 | 32.5 |
| | | Drunkard's Odometry | 74.2 | 1.54 | 85.7 | 7.32 | |
| 4 | 0 | RAFT-3D | 83.3 | 1.62 | 86.8 | 9.58 | 28.0 |
| | | Drunkard's Odometry | 79.3 | 2.09 | 83.6 | 12.27 | |
| | 1 | RAFT-3D | 83.4 | 1.36 | 86.8 | 9.20 | 31.6 |
| | | Drunkard's Odometry | 79.2 | 1.80 | 83.2 | 12.11 | |
| | 2 | RAFT-3D | 81.6 | 1.42 | 84.8 | 9.76 | 31.5 |
| | | Drunkard's Odometry | 76.6 | 1.87 | 80.3 | 12.82 | |
| | 3 | RAFT-3D | 77.4 | 1.77 | 81.6 | 11.13 | 32.43 |
| | | Drunkard's Odometry | 71.0 | 2.30 | 75.7 | 14.74 | |
| 5 | 0 | RAFT-3D | 85.7 | 1.16 | 89.1 | 8.82 | 27.7 |
| | | Drunkard's Odometry | 82.3 | 1.53 | 86.1 | 11.27 | |
| | 1 | RAFT-3D | 85.4 | 1.15 | 87.6 | 9.08 | 25.1 |
| | | Drunkard's Odometry | 81.7 | 1.49 | 84.9 | 11.37 | |
| | 2 | RAFT-3D | 82.1 | 1.29 | 84.9 | 10.26 | 26.1 |
| | | Drunkard's Odometry | 77.1 | 1.65 | 80.3 | 12.94 | |
| | 3 | RAFT-3D | 75.7 | 1.76 | 79.9 | 12.09 | 30.21 |
| | | Drunkard's Odometry | 69.19 | 2.19 | 74.2 | 15.74 | |

Table B.1: **2D and 3D flow metrics** in each of the four levels of difficulty of the three test scenes of the Drunkard's Dataset for RAFT-3D and our Drunkard's Odometry.

We observed that our model can be trained at a faster learning rate ($2 \times 10^{-4}$) than RAFT-3D ($10^{-4}$), possible due to the additional supervision. Our models need optical flow and depth supervision, in addition to pose, for convergence (without all these, the training diverges eventually). Inverting the order of the input image pair during training, as a data augmentation technique, has a significant impact in reducing the translation pose estimation error (notice how the translation curve keeps pointing down even in the last training stages, which does not happen without such augmentation).

## D  Architecture Details

The architectural details for the feature encoder, the context encoder, the pose network and the update block of our network are shown in Figures D.1 and D.2. The figures show network blocks (in gray) and input, intermediate and output variables (in blue, green and red respectively). The pose network is a modification of the one used in ManyDepth (3) (Copyright © Niantic, Inc. 2021. See project website for more license details), so the initial guess of the relative camera pose given by the pose network is in Lie algebra ($\mathbf{t}_c^{\text{pre}}$) and it is exponential mapped back to $\text{SE}(3)$ ($\mathbf{T}_c^{\text{pre}}$) to initialize $\mathbf{T}_c$ and $\mathbf{T}$ before the iterative block. In the update block, the input pose twist $\mathbf{t}_c^k$ is cloned to every pixel value of a tensor with the same resolution as the other input variables ($H/8 \times W/8$) to be able to stack them together.

## References

[1] S. K. Ramakrishnan, A. Gokaslan, E. Wijmans, O. Maksymets, A. Clegg, J. Turner, E. Undersander, W. Galuba, A. Westbury, A. X. Chang, *et al.*, "Habitat-matterport 3d dataset (hm3d): 1000 large-scale 3d environments for embodied ai," *arXiv preprint arXiv:2109.08238*, 2021.

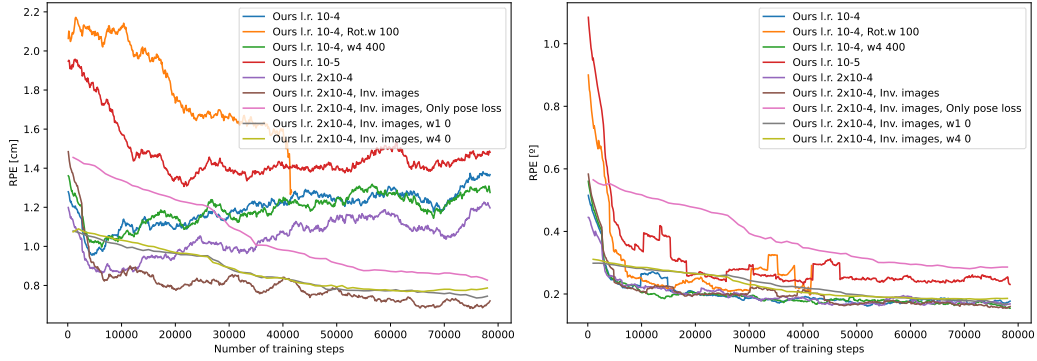

Figure C.1: **Ablations and hyperparameter influence on model training.** We show the relative position errors (RPE) in the three Drunkard's Dataset test scenes for difficulty level 1 and for different model variations of our Drunkard's Odometry during training. With "l.r." we denote different learning rates. If applied, data augmentation with inverted images is shown as "inv. images" and different loss term weights are shown as "w1", "w4", "Rot. w" followed by their values.

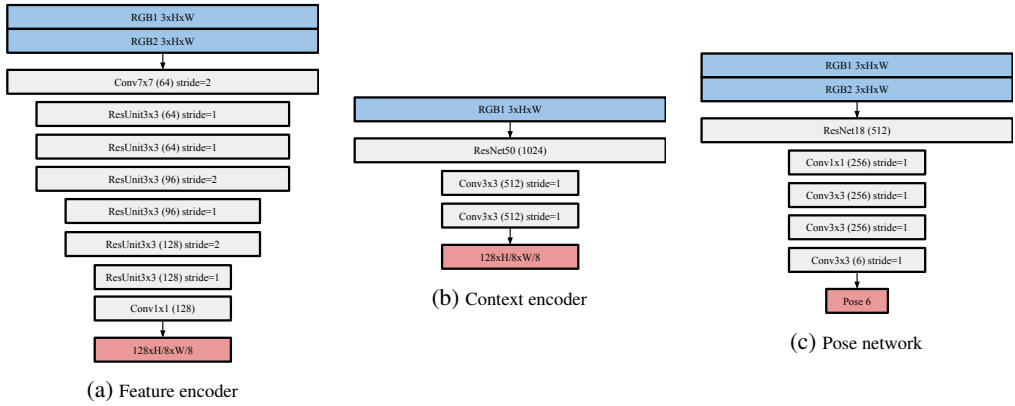

(a) Feature encoder

(b) Context encoder

(c) Pose network

Figure D.1: **Network details** for the feature encoder, the context encoder and the pose network.

[2] Z. Teed and J. Deng, "Raft-3d: Scene flow using rigid-motion embeddings," in *Proceedings of the IEEE/CVF Conference on Computer Vision and Pattern Recognition*, pp. 8375–8384, 2021.

[3] J. Watson, O. Mac Aodha, V. Prisacariu, G. Brostow, and M. Firman, "The temporal opportunist: Self-supervised multi-frame monocular depth," in *Proceedings of the IEEE/CVF Conference on Computer Vision and Pattern Recognition*, pp. 1164–1174, 2021.

4

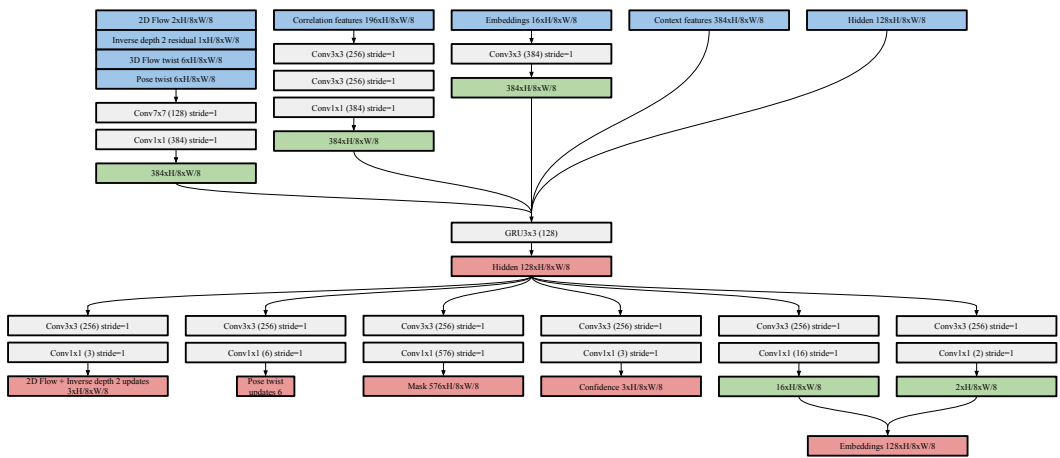

Figure D.2: **Network details for the update block.**


\fi

\end{document}